\documentclass[sigconf]{acmart}
\usepackage[latin9]{inputenc}
\usepackage{color}
\usepackage{array}
\usepackage{multirow}
\usepackage{amsbsy}
\usepackage{graphicx}
\makeatletter

\providecommand{\tabularnewline}{\\}
\AtBeginDocument{%
  \providecommand\BibTeX{{%
    \normalfont B\kern-0.5em{\scshape i\kern-0.25em b}\kern-0.8em\TeX}}}

\copyrightyear{2021} 
\acmYear{2021} 
\setcopyright{acmcopyright}
\acmConference[MM '21]{Proceedings of the 29th ACM
International Conference on Multimedia}{October 20--24, 2021}{Virtual Event, China}
\acmBooktitle{Proceedings of the 29th ACM International Conference on Multimedia
(MM '21), October 20--24, 2021, Virtual Event, China}
\acmPrice{15.00}
\acmDOI{10.1145/3474085.3475499}
\acmISBN{978-1-4503-8651-7/21/10}

\makeatother

\begin{document}
\fancyhead{} 
\title{Domain-Aware SE Network for Sketch-based Image Retrieval with Multiplicative
Euclidean Margin Softmax}

\author{Peng Lu}
\affiliation{
    \institution{Shenzhen International Graduate School, Tsinghua University}
    \city{Shenzhen}  \country{China}  
}
\email{lup21@mails.tsinghua.edu.cn}

\author{Gao Huang}
\affiliation{
    \institution{Department of Automation, Tsinghua University}
    \city{Beijing}  \country{China}  
}
\email{gaohuang@tsinghua.edu.cn}

\author{Hangyu Lin}
\affiliation{
    \institution{School of Data Science, Fudan University} 
    \city{Shanghai} \country{China} 
}
\email{linhy960303@gmail.com}

\author{Wenming Yang}
\authornote{indicates corresponding author.}
\affiliation{
    \institution{Shenzhen International Graduate School \& Department of Electronic Engineering, Tsinghua University}
    \city{Shenzhen}  \country{China}  
}
\email{yang.wenming@sz.tsinghua.edu.cn}

\author{Guodong Guo}
\affiliation{
    \institution{IDL, Baidu Research}
    \institution{National Engineering Lab for Deep Learning Technology and Application}
    \city{Beijing}  \country{China}  
}
\email{guoguodong01@baidu.com}

\author{Yanwei Fu}
\authornote{Yanwei Fu is with the School of Data Science, MOE Frontiers Center for Brain Science, and Shanghai Key Lab of Intelligent Information Processing, Fudan University.}
\affiliation{
    \institution{School of Data Science, Fudan University} 
    \city{Shanghai} \country{China} 
}
\email{yanweifu@fudan.edu.cn}

\begin{abstract}
This paper proposes a novel approach for Sketch-Based Image Retrieval (SBIR), for which the key is to bridge the gap between sketches and photos in terms of the data representation. Inspired by channel-wise attention explored in recent years, we present a Domain-Aware Squeeze-and-Excitation (DASE) network, which seamlessly incorporates the prior knowledge of sample sketch or photo into SE module and make the SE module capable of emphasizing appropriate channels according to domain signal. Accordingly, the proposed network can switch its mode to achieve a better domain feature with lower intra-class discrepancy. 
Moreover, while previous works simply focus on minimizing intra-class distance and maximizing inter-class distance, 
we introduce a loss function, named Multiplicative Euclidean Margin Softmax (MEMS), which introduce multiplicative Euclidean margin into feature space and ensure that \emph{the maximum intra-class distance is smaller than the minimum inter-class distance}. This facilitates learning a highly discriminative feature space and ensures a more accurate image retrieval result. Extensive experiments are conducted on two widely used SBIR benchmark datasets. Our approach achieves better results on both datasets, surpassing the state-of-the-art methods by a large margin. 
\end{abstract}


\maketitle

\section{Introduction}

\begin{figure}
\centering \includegraphics[width=7cm]{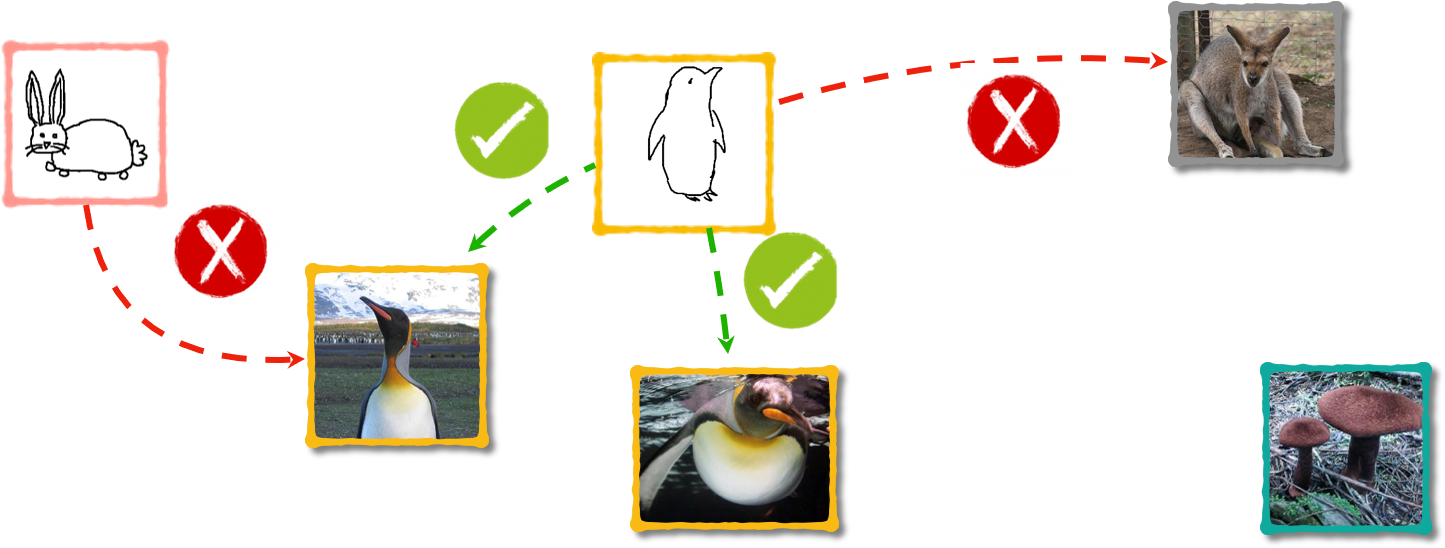}
\caption{\label{fig:Illustration-of-SBIR}Illustration of SBIR task.}
\vspace{-0.1cm}
\end{figure}

Touch-screen devices, such as smartphones and iPad, enable users to
draw free-hand sketches conveniently. These sketches are highly iconic,
succinct, and abstract representations, and usually convey richer
and more accurate information than texts in various scenarios. Consequently,
many novel applications related with sketch have sprung up. One representative
example is the Sketch-Based Image Retrieval (SBIR), which has attracted
great attention from the computer vision community during the past
decades\ \citep{kato1992sketch,yu2016sketch,song2017deep,liu2017deep,zhang2018generative,shen2018zero,liu2017deep}.
For the SBIR task, learning good representations for both sketches
and photos is vital and remains a challenging problem\ \citep{yu2016sketch,yu2017sketch,song2017deep,saavedra2014sketch,saavedra2015sketch,wang2015sketch}.

Given a query sketch, the objective of SBIR is to find relevant photos that are semantically related, \emph{e.g.}, they come from the same category (as shown in Figure\ \ref{fig:Illustration-of-SBIR}). This task appears to be easy for humans but is difficult for a machine. The main challenge comes from the fact that there is a huge gap between the data representation in the two domains: the sketches are represented by highly iconic, abstract, and sparse lines, while the photos are composed of dense color pixels with rich texture information. This domain gap obstructs the SBIR model in exploiting the shared semantics and discriminative representations for both photos and sketches.

Recently, plenty of works have been proposed to address this problem.
A popular approach is constructing an excellent intermediate representation,
\textit{i.e.}, converting photos to edge maps\ \citep{yu2016sketch,song2017deep,liu2017deep,zhang2018generative}
or translating sketches into the photo domain using generative models
\citep{zhang2018generative}. Another widely adopted approach is learning
a semi-heterogeneous network in an end-to-end manner\ \citep{shen2018zero,liu2017deep}.
These approaches have either large computation consumption or redundant
parameters. On the contrary, TC-net\ \citep{lin2019tc} uses a single
CNN to process both sketches and photos. High retrieval accuracy is
achieved with this scheme. However, vanilla CNN is domain-agnostic
and thus can not fully utilize the prior knowledge about which domain
the sample comes from.

Therefore, we propose \textcolor{black}{a Domain-Aware Squeeze-and-Excitation
(DASE) network}, which can embed sketches and photos respectively
according to their domain with a negligible increase of computation
and parameter. This function is achieved via DASE module. In vanilla
SE module\ \citep{hu2017squeeze}, the convolutional features are firstly
squeezed into a low dimensional embedding by an encoder network, and
then they are decoded to generate a channel-wise attention vector
which is applied to the original feature maps. In DASE module, we
simply append one binary-value code that indicates which domain
the sample comes from, to the low dimensional embedding. With this
modification, DASE module can emphasize different channels according
to the domain knowledge. Extensive experiments prove that this change
is simple yet highly effective.

Further, loss function is essential in learning a discriminative feature
space. Specifically, to achieve high retrieval accuracy, an ideal feature criterion is expected where the maximum intra-class distance is less than the minimum inter-class distance. However, previous works on SBIR do not optimize networks toward this end. 
Cross-view Pairwise Loss\ \citep{liu2017deep} and Semantic Loss\ \citep{zhang2018generative} are utilized to minimize intra-class distance and maximize inter-class distance in binary space. But they do not introduce any margin and guarantee the above ideal feature criterion. Moreover, these losses are difficult to optimize since they are non-convex and non-smooth, and optimization algorithms based on alternating iteration is required, which are very cumbersome.

In this work, we propose maximizing the inter-class distance to be several times the intra-class distance in Euclidean space. This enables us to derive a novel loss function -- Multiplicative Euclidean Margin Softmax (MEMS), which controls the ratio of the intra-class and the inter-class distance jointly. With appropriate ratio, the maximum intra-class distance is less than the minimum inter-class distance.
This loss can be easily optimized using back-propagation and can generate a discriminative feature space. Further, we provide theoretical analysis to explain how the MEMS loss works and give a lower bound of the multiplicative margin.
Experiments show that the MEMS loss can yield highly accurate retrieval results, which surpass all existing algorithms by a large margin. 
The results generated by MEMS loss can be kept in a hashing scheme that converts learned features into low dimensional binary codes. This post-processing step can be realized with a spectral normalized perception which is also easy to optimize.
The experimental results validate the effectiveness of our model.

\vspace{0.05in}

\noindent \textbf{Contributions}. The main contributions of this paper include: (1) A novel network architecture DASE is introduced to incorporate the additional information about the domain attribution, which boosts the retrieval performance with both shallow and deep networks; (2) A novel loss function, the Multiplicative Euclidean Margin Softmax (MEMS), that optimizes a discriminative feature space where the maximum intra-class distance is smaller than the minimum inter-clsss distance; (3) Theoretical analysis is provided on the properties of the MEMS loss; (4) New state-of-the-art results have been obtained on several competitive SBIR tasks.

\section{Related Work}

\noindent \textbf{SBIR.} Sketch-Based Image Retrieval (SBIR) aims
to retrieve similar semantic meaning images as the query sketch. A
typical solution is to learn a shared embedding space for both sketches
and images. Such a common space facilitates the ranking of similarity
of sketches and images. Previous methods transfer photos into sketch-tokens
and then extract hand-craft sketch features\ \citep{lowe1999object,dalal2005histograms,hu2010gradient,saavedra2014sketch}
or deep features\ \citep{wang2015sketch,yu2016sketch,yu2017sketch,song2017deep,liu2017deep}
to represent the sketches. Recent deep learning based architectures
\citep{zhang2018generative,lin2019tc,pandey2020stacked,li2019bi}
enable cross-domain learning in an end-to-end manner. To accelerate
the retrieval in a large-scale dataset, hashing based models\ \citep{liu2017deep,shen2018zero,zhang2018generative,yelamarthi2018zero}
have also been studied.

\vspace{0.05in}

\noindent \textbf{Attention in CNNs.} Attention mechanism enables
networks to assign different significance on different parts of the
input. The important feature expressions are amplified while the less
useful ones are suppressed. To better apprehend digital images, many
convolutional networks with spatial attention\ \citep{jaderberg2015spatial,dai2017deformable,wang2018non,li2019selective}
and channel-wise attention\ \citep{hu2017squeeze,cao2019gcnet} are
proposed to emphasize the informative region or channels. SENet\ \citep{hu2017squeeze}
proposes a squeeze-and-excitation structure. Global context is aggregated
by global average pooling for each channel. The aggregated information
is then processed and used to reallocate attention over channels by
feature fusion. To tackle the SBIR problem, spatial attention\ \citep{song2017deep,sketchanet2017ijcv,zhang2018generative}
has been utilized. However, channel-wise attention has never been
explored in cross-modal embedding for sketches and photos so far.

\vspace{0.05in}

\noindent \textbf{Loss functions.} Many metric learning based methods
\citep{hu2014discriminative,lu2015multi,wen2016discriminative,wang2014learning}
proposed learning deep features by the loss functions with Euclidean
distances. To make the learned features more discriminative, other
variants of loss functions have been investigated recently, such as
the contrastive loss\ \citep{chopra2005learning,hadsell2006dimensionality},
triplet loss\ \citep{schroff2015facenet}, and softmax-based losses\citep{liu2016large,liu2017sphereface,wang2018cosface,deng2019arcface}.
Particularly, the contrastive and triplet losses aim at increasing
the Euclidean margin in an additive manner, which might be neglected
as there is no upper bound for Euclidean distance. A-softmax\ \citep{liu2017sphereface}
optimize the inter-class angular distance to be several times the
intra-class angular distance to guarantee that the least inter-class
distance is more than the largest intra-class distance. Furthermore,
prototypical loss\ \citep{snell2017prototypical} is also a variant
of softmax which incorporates the Euclidean distance.

\begin{figure*}[t]
\centering{}\includegraphics[width=17cm]{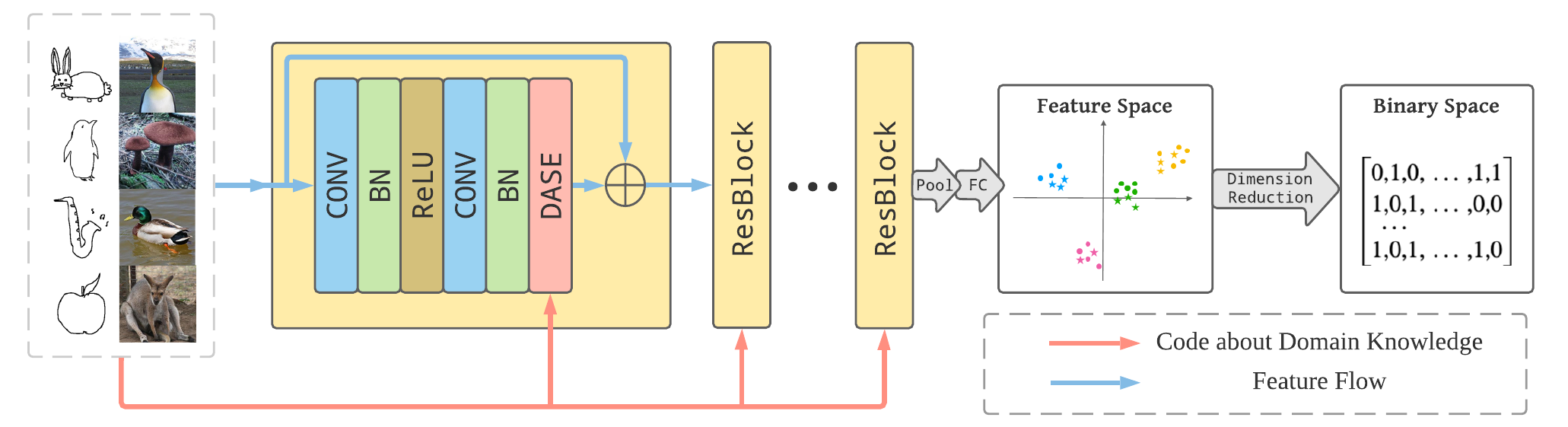}\caption{\label{fig:framework}Illustration of our framework. We mix sketches and photos as input of the DASE network, to which the proposed DASE module is added. Domain knowledge is encoded in DASE modules in each ResBlock. After training, the output features or binarized code can be used to conduct SBIR, where Euclidean distance and Hamming distance are used as metric, respectively.}
\end{figure*}

\section{Method}

In this section, we present the proposed DASE network and the MEMS
loss for SBIR task in detail. The SBIR task is formulated as\ \citep{liu2017deep,zhang2018generative}.
We have the realistic photo set $\mathcal{P}=\left\{ \left(p_{i},y_{_{i}}\right)\mid y_{i}\in\mathcal{Y}\right\} $,
and the sketch set $\mathcal{S}=\left\{ \left(s_{j},y_{j}\right)\mid y_{j}\in\mathcal{Y}\right\} $
respectively. Given a query sketch, the objective of SBIR is to find
all of its relevant photos that semantically belong to the same category
as the sketch. As the setting defined in\ \citep{liu2017deep}, the
same photo set $\mathcal{P}$ is used for both training and test,
as the retrieval galleries. Sketch set $\mathcal{S}$ is split into
train and test sets, with the same label set $\mathcal{Y}$.

The feature representation networks of sketches and photos are denoted
as $\mathcal{F}_{p}$ and $\mathcal{F}_{s}$ respectively. The two
networks project sketches and photo to a shared space, \textit{i.e.}
$\mathcal{F}_{p}(p_{i}),\mathcal{F}_{s}(s_{j})\in\mathcal{X}$;
thus we have the set $\mathcal{D}=\left\{ \left(\boldsymbol{x},y_{\boldsymbol{x}}\right)\vert y_{\boldsymbol{x}}\in\mathcal{Y}\right\} $,
where the feature $\boldsymbol{x}\in\mathcal{X}$ can come either
from photo domain or sketch domain and the label $y_{\boldsymbol{x}}$
is identical with the source sample of feature $\boldsymbol{x}$.
We further introduce the denotation, $\mathcal{D}_{y}=\left\{ \boldsymbol{x}\mid y_{\boldsymbol{x}}=y\right\} $
for convenience.
\vspace{-0.1in}

\subsection{\label{subsec:Overview}Overview}

To facilitate learning feature representations for sketches and photos,
the proposed Domain-Aware SE (DASE) network is shared by both domains.
This network is composed of several ResBlocks in which the DASE module
is inserted (see in Figure\ \ref{fig:framework}). In contrast to
vanilla SE module\ \citep{hu2017squeeze}, DASE module receives code
about the domain of input image and emphasizes different channels
individually. Further, to bridge the data representation gap between
the sketch and photo domains, we present a novel loss function --
Multiplicative Euclidean Margin Softmax (MEMS) loss, which can efficiently
learn a unified embedding space. The MEMS loss optimizes the larger
inter-class and smaller intra-class distance over the photo and sketch
set. In other words, the instances of sketches and photos in the same/different
class should be close/far from each other.
\vspace{-0.1in}

\subsection{\label{subsec:DASE}Domain-Aware SE Network}

\begin{figure}
\centering{}\includegraphics[scale=0.32]{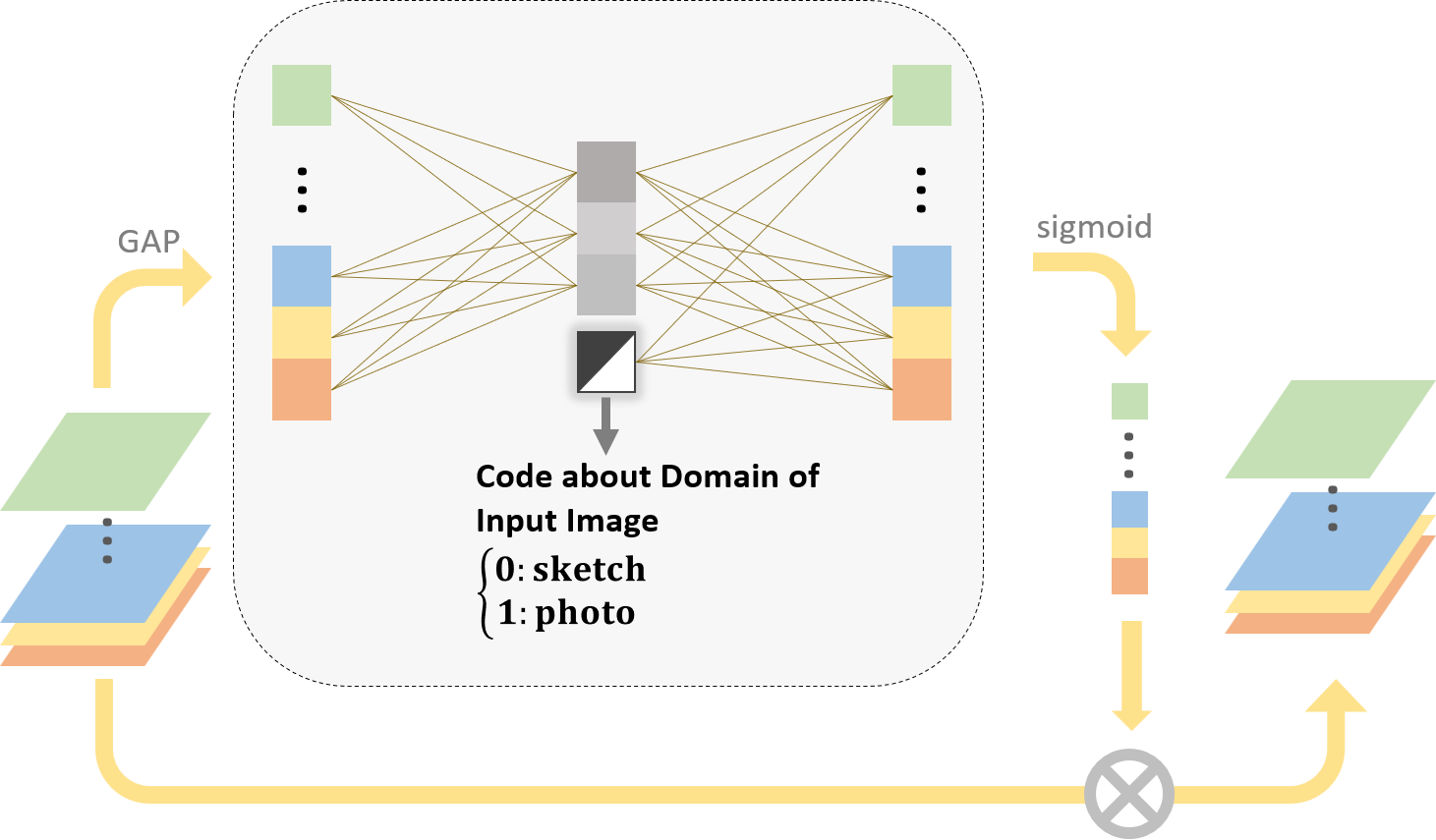} 
\caption{\label{fig:DASE}The structure of DASE module. The activation function for the intermediate FC layer is Sigmoid.}
\vspace{-0.1in}
\end{figure}

Learning common feature representations for sketch and photo domains
is a non-trivial task. Obviously, sketches are iconic and abstract with various deformation levels; while photos are realistic images with rich color, texture, and shape information.

Rather than using independent sub-networks to process the photos and sketches separately, our model learns a single network to analyze the photos and sketches jointly. Such a SiameseNet-like network is inspired by the fact that SiameseNet is efficient in learning the embedding space across different domains (\emph{e.g.}, image-text embedding\ \citep{siamense_image_text}, or person re-identification\ \citep{gated_siamese_eccv2016}). The semi-heterogeneous network is proposed in\ \citep{liu2017deep}, including a shared Siamese sub-network and separate sub-networks for each domain.

In contrast to\ \citep{liu2017deep}, we further propose Domain-Aware Squeeze-and-Excitation (DASE) module, which is shown in Figure\ \ref{fig:DASE}, to learn an effective feature representation. Particularly, our DASE module is an extension of SE module\ \citep{hu2017squeeze} which provides an explicit mechanism to re-weight the importance of channels after each block in the network. As shown in Figure\ \ref{fig:DASE}, our DASE module utilizes an encoder-decoder structure followed by a sigmoid activation. Within the intermediate space, a \emph{binary code} is added to indicate whether the input image is a sketch or a photo. The outputs of sigmoid activation are the feature attention vector over channels. This conditional structure can thus help capture different characteristics of input images conditioned on which domain they come from.

With the DASE module, our feature extractor, which is a Siamese Network with the domain-conditioned structure, is learned over two domains to explore shared semantics in a common feature space. The binary code serves as a controlling gate and endows the flexibility of network in learning from each individual domain.

\subsection{\label{subsec:MEMS}Multiplicative Euclidean Margin Softmax Loss}

The loss function is vital in successfully optimizing the networks
over the recognition tasks, especially in our cross-domain scenario.
To embed the photo and sketch into a shared space, we propose the
Multiplicative Euclidean Margin Softmax loss, which exploits the strategy
of maximizing the intra-class distance and minimizing the inter-class
distance.

Given the category $y\in\mathcal{Y}$, the maximum intra-class distance
can be defined by $\max_{\boldsymbol{x},\boldsymbol{x}'\in\mathcal{D}_{y}}\mathrm{d}\left(\boldsymbol{x},\boldsymbol{x}'\right)$
and the minimum inter-class distance would be $\min_{\boldsymbol{x}\in\mathcal{D}_{y},\boldsymbol{x}'\in\left(\bigcup_{y'\ne y}\mathcal{D}_{y'}\right)}\mathrm{d}\left(\boldsymbol{x},\boldsymbol{x}'\right)$,
where $\mathrm{d}\left(\cdot,\cdot\right)$ can be any kind of differentiable
distance metric. Thus to develop our formulation, we optimize our
framework by enforcing that the maximum intra-class distance to be
smaller than minimum inter-class distance,

\begin{equation}
\forall y\in\mathcal{Y},\;\max_{\boldsymbol{x},\boldsymbol{x}'\in\mathcal{D}_{y}}\mathrm{d}\left(\boldsymbol{x},\boldsymbol{x}'\right)\le\min_{\boldsymbol{x}\in\mathcal{D}_{y},\boldsymbol{x}'\in\left(\bigcup_{y'\ne y}\mathcal{D}_{y'}\right)}\mathrm{d}\left(\boldsymbol{x},\boldsymbol{x}'\right).\label{eq:target}
\end{equation}

Rather than directly optimizing Eq (\ref{eq:target}) over all instances,
we compute the prototypes $\left\{ \boldsymbol{c}_{y}\right\} _{y\in\mathcal{Y}}\subset\mathcal{X}$
of each class, and characterize the distribution of instances in feature
space. If MEMS loss is well optimized, instances will be closer to
their corresponding prototype than other prototypes in feature space.
This is as, 
\begin{equation}
\forall\left(\boldsymbol{x},y_{\boldsymbol{x}}\right)\in\mathcal{D},\forall y\in\mathcal{Y}\wedge y\ne y_{\boldsymbol{x}},m\cdot\mathrm{d}\left(\boldsymbol{x},\boldsymbol{c}_{y_{\boldsymbol{x}}}\right)\le\mathrm{d}\left(\boldsymbol{x},\boldsymbol{c}_{y}\right),\label{eq:multiplicative-euclidean-distance}
\end{equation}
where $m\ge1$ is referred to as the margin constant.

For convenience, we denote $\mathcal{R}_{y,y'}$ as a region in the
feature space, 
\[
\boldsymbol{x}\in\mathcal{R}_{y,y'}\;\Leftrightarrow\;m\cdot\mathrm{d}\left(\boldsymbol{x},\boldsymbol{c}_{y}\right)\le\mathrm{d}\left(\boldsymbol{x},\boldsymbol{c}_{y'}\right).
\]
Also, we denote $\mathcal{R}_{y}$ as a region such that 
\[
\boldsymbol{x}\in\mathcal{R}_{y}\;\Leftrightarrow\;\forall y'\in\mathcal{Y}\wedge y'\ne y,m\cdot\mathrm{d}\left(\boldsymbol{x},\boldsymbol{c}_{y}\right)\le\mathrm{d}\left(\boldsymbol{x},\boldsymbol{c}_{y'}\right),
\]
which takes account different data distributions of each class. It
is easy to prove that $\mathcal{R}_{y}=\bigcap_{y'\ne y}\mathcal{R}_{y,y'}$.
Note that if Eq (\ref{eq:multiplicative-euclidean-distance}) holds,
\[
\forall\left(\boldsymbol{x},y_{\boldsymbol{x}}\right)\in\mathcal{D},\;\boldsymbol{x}\in\mathcal{R}_{y_{\boldsymbol{x}}},
\]
and thus we can derive a sufficient condition for Eq. (\ref{eq:target}),
\begin{equation}
\forall y\in\mathcal{Y},\max_{\boldsymbol{x},\boldsymbol{x}'\in\mathcal{R}_{y}}\mathrm{d}\left(\boldsymbol{x},\boldsymbol{x}'\right)\le\min_{\boldsymbol{x}\in\mathcal{R}_{y},\boldsymbol{x}'\in\left(\bigcup_{y'\ne y}\mathcal{R}_{y'}\right)}\mathrm{d}\left(\boldsymbol{x},\boldsymbol{x}'\right).\label{eq:target-sufficient}
\end{equation}

\noindent \textbf{MEMS loss.} To incorporate this understanding and
optimize Eq (\ref{eq:multiplicative-euclidean-distance}) by softmax
loss, this gives us a novel loss function-- Multiplicative Euclidean
Margin Softmax (MEMS) loss:

\begin{equation}
\mathcal{L}_{ems}=\frac{1}{N}\sum_{i=1}^{N}-\log\frac{e^{-m^{2}\left\Vert \boldsymbol{x}_{i}-\boldsymbol{c}_{y_{i}}\right\Vert _{2}^{2}}}{e^{-m^{2}\left\Vert \boldsymbol{x}_{i}-\boldsymbol{c}_{y_{i}}\right\Vert _{2}^{2}}+\sum_{j\ne y_{i}}e^{-\left\Vert \boldsymbol{x}_{i}-\boldsymbol{c}_{j}\right\Vert _{2}^{2}}},\label{eq:mems-function}
\end{equation}

\noindent where $\boldsymbol{x}_{i}$ indicates the feature extracted
by the last layer of feature representation network. $\boldsymbol{c}_{j}$
is the center of $j$-th category. We take the center $\boldsymbol{c}_{j}$
as the parameters, and update $\boldsymbol{c}_{j}$ dynamically, rather
than directly using the average feature center. In Eq (\ref{eq:mems-function}),
we employ the negative squared Euclidean distance to measure the confidence
of $x_{i}$ being $\left(-\left\Vert \boldsymbol{x}_{i}-\boldsymbol{c}_{j}\right\Vert _{2}^{2}\right)$.
Particularly, in binary classification, $x_{i}$ is labelled as class
$1$ if $m\left\Vert \boldsymbol{x}-\boldsymbol{c}_{1}\right\Vert _{2}<\left\Vert \boldsymbol{x}-\boldsymbol{c}_{2}\right\Vert _{2}$,
and otherwise, as class 2.

\subsection{Theoretical analysis and lower bound of $m$}

\noindent The property of MEMS is largely determined by the value
of $m$. Intuitively, a larger $m$ makes decision boundaries closer
to corresponding prototypes and the distribution of features more
compact. This produces a more discriminative metric space. However,
overly large $m$ tends to make the training process unstable due
to the inherent variances among samples in each category.

Therefore, it is necessary to find the minimum $m$ to ensure that,
for every sample, and in metric space, the maximum intra-class distance
is smaller than minimum inter-class distance. We give some theoretical
analysis about the Eq (\ref{eq:multiplicative-euclidean-distance})
and Eq (\ref{eq:target-sufficient}). Most importantly, if we adopt
Euclidean distance as $\mathrm{d}\left(\cdot,\cdot\right)$, $m\ge2+\sqrt{3}$
is the sufficient and necessary condition for Eq (\ref{eq:target-sufficient})
given Eq (\ref{eq:multiplicative-euclidean-distance}). We show a
brief proof below and the details can be found in Appendix.

\noindent \textbf{Sufficient. }If $\mathrm{d}(\boldsymbol{x},\boldsymbol{x}')=\left\Vert \boldsymbol{x}-\boldsymbol{x}'\right\Vert _{2}$,
$\mathcal{R}_{y,y'}$ is a $n$-ball (ball in $n$-dimensional space)
with the center $\left(\boldsymbol{c}_{y}+\left(\boldsymbol{c}_{y}-\boldsymbol{c}_{y'}\right)/\left(m^{2}-1\right)\right)$
and the radius $\left(m/m^{2}-1\right)\left\Vert \boldsymbol{c}_{y}-\boldsymbol{c}_{y'}\right\Vert _{2}$.
Thus in binary categorization, the minimum value of $m$ is $2+\sqrt{3}$.
With the growth of the number of categories, the minimum value is
reduced monotonously. So it is sufficient to have $m\ge2+\sqrt{3}$
guaranteeing perfect discrimination.

\noindent \textbf{Necessary. }We illustrate the necessity in multi-class
cases, since, if two prototypes are far from the others, their relationship
will resemble the one in binary case. So $m\ge2+\sqrt{3}$ is necessary
regardless of the number of categories.

\subsection{Differences with other losses}

We further discuss the differences between our MEMS loss and other
losses.

\noindent \textbf{Prototypical loss}\ \citep{snell2017prototypical}
It is defined as
\vspace{-0.1in}
\begin{equation}
\mathcal{L}_{proto}=\frac{1}{N}\sum_{i}^{N}-\log\frac{e^{-\left\Vert \boldsymbol{x}_{i}-\boldsymbol{c}_{y_{i}}\right\Vert _{2}^{2}}}{\sum_{j}e^{-\left\Vert \boldsymbol{x}_{i}-\boldsymbol{c}_{j}\right\Vert _{2}^{2}}}.\label{eq:prototypical-function}
\vspace{-0.1in}
\end{equation}

\noindent The prototypical loss is used in one-shot classification
where only a few training instances are available for each class.
Thus $\boldsymbol{c}_{j}$ is directly computed as the averaged mean
of training instances; in contrast, our MEMS loss is a generalized
softmax loss, and optimizes $\boldsymbol{c}_{j}$ via back-propagation.
Furthermore, the prototypical loss is a special case of MEMS loss when the margin is 1. It is sufficient for classification task and insufficient for training a discriminative feature space for retrieval task. Therefore a margin constant $m$ is introduced to make a balance between enlarging the inter-class distance and shrinking the intra-class distance.

\noindent \textbf{Angular Margin loss}. We further discuss the difference
between MEMS loss and angular margin loss. Similar to Euclidean distance,
angular distance based loss functions, such as A-Softmax\ \citep{liu2017sphereface}
and LMCL\ \citep{wang2018cosface}, are also employed in learning a
shared space in many tasks, \emph{e.g.}, face recognition. These angular
margin losses aim at learning a discriminative distribution on a hypersphere.
As shown in Table\ \ref{tab:Sphere-and-cosine}, they define different
similarity functions for the instances of different classes. Note
that $\psi(\theta)$ is an artificial piece-wise function that serves
as the extension of $\cos\left(m\cdot\theta\right)$ , to overcome
its non-monotonicity. Nevertheless, it is non-trivial to define the
$\psi(\theta)$ in A-softmax function as stated in\ \citep{wang2018cosface}.
The scalar $s$ in LMCL is used to expand the range of similarity
function; otherwise, the output of softmax function would be closed
to the uniform distribution over all categories.

\begin{table}[H]
\begin{centering}
\begin{tabular}{c|c|c}
\hline 
Similarity  & intra-class  & inter-class ($j\ne y_{i}$)\tabularnewline
\hline 
\hline 
A-Softmax  & $\left\Vert \boldsymbol{x}_{i}\right\Vert _{2}\cdot\psi(\theta_{y_{i}})$  & $\left\Vert \boldsymbol{x}_{i}\right\Vert _{2}\cdot\cos(\theta_{j})$\tabularnewline
\hline 
LMCL  & $s\cdot\left(\cos(\theta_{y_{i}})-m\right)$  & $s\cdot\cos(\theta_{j})$\tabularnewline
\hline 
MEMS (ours)  & $-m^{2}\left\Vert \boldsymbol{x}_{i}-\boldsymbol{c}_{y_{i}}\right\Vert _{2}^{2}$  & $-\left\Vert \boldsymbol{x}_{i}-\boldsymbol{c}_{j}\right\Vert _{2}^{2}$\tabularnewline
\hline 
\end{tabular}
\par\end{centering}
\begin{centering}
\vspace{0.1in}
 
\par\end{centering}
\caption{\label{tab:Sphere-and-cosine}The similarity functions of A-Softmax,
LMCL and MEMS loss. $\theta_{y}$ represents angular distance between
the feature vector and the prototype of category $y$.}
\end{table}

\noindent \textbf{Triplet loss}\ \citep{schroff2015facenet} It is
defined as

\begin{equation}
\mathcal{L}_{tri}=\sum_{i}^{N}\left[\left\Vert \boldsymbol{x}_{i}^{a}-\boldsymbol{x}_{i}^{p}\right\Vert _{2}^{2}-\left\Vert \boldsymbol{x}_{i}^{a}-\boldsymbol{x}_{i}^{n}\right\Vert _{2}^{2}+\alpha\right]_{+},\label{eq:triplet-loss}
\end{equation}
 where $\boldsymbol{x}_{i}^{a}$ and $\boldsymbol{x}_{i}^{p}$ share
same class while $\boldsymbol{x}_{i}^{a}$ and $\boldsymbol{x}_{i}^{n}$
come from different class. $\alpha$ is the margin between the distance
of positive pairs and negative pairs. Different from the proposed
MEMS loss, this margin is additive and can not ensure that the \textit{maximum}
intra-class distance is smaller than the \textit{minimum} inter-class
distance. In addition, triplet loss computes the distance between
two instance pairs, and its sampling complexity is $\mathcal{O}\left(N^{3}\right)$.
By contrast, the sampling complexity of MEMS loss is s $\mathcal{O}\left(N\right)$
as it computes distance between each instance and prototypes.

\section{Experiments}

\subsection{\label{subsec:Datasets-and-Settings}Datasets and Settings}

\noindent \textbf{Datasets.} Our model is evaluated on two large-scale
sketch-photo datasets: TU-Berlin\ \citep{eitz2012humans} Extension
and Sketchy\ \citep{sangkloy2016sketchy} Extension. The former includes
20,000 sketches uniformly distributed among 250 categories. Additionally,
204,489 natural images provided in\ \citep{zhang2016sketchnet} are
utilized as the photo gallery. The Sketchy database consists of 75,471
hand-drawn sketches and 12,500 corresponding photos from 125 categories.
It was extended by another 60,502 photos for SBIR task in\ \citep{liu2017deep}.
Following the settings in\ \citep{liu2017deep,zhang2018generative},
10/50 sketches from each category are picked as the query set for
TU-Berlin/Sketchy dataset, and the rest are used for training. All
gallery photos are used in both training and testing phases.

\noindent \textbf{Implementation.} Our method is implemented using
Pytorch with single 1080Ti GPU. We use Adam optimizer\ \citep{kingma2014adam}
with parameters $\beta_{1}=0.9,\beta_{2}=0.999,\lambda=0.0001$. The
learning rate is set to $0.0001$ and linearly decays to $0$ during
the second half of training. The model converges after training for
200k iterations. We use the backbone network of ResNeXt-101, which
is composed of 33 ResNeXt Blocks. The proposed DASE module
is added in each block. We use $m=4$ in the proposed MEMS loss. Our code is available at: {\color{blue}{\href{https://github.com/Ben-Louis/SBIR-DASE-MEMS}{https://github.com/Ben-Louis/SBIR-DASE-MEMS}}}

\noindent \textbf{SBIR hashing. }SBIR can be conducted using either
real-value or binary-value vectors as features. The latter is named
\textit{SBIR hashing}, which greatly accelerates the speed of SBIR
tasks but requires auxiliary design or process. The proposed method
can generate both styles of features. Particularly, for SBIR hashing,
we utilize a simple hashing scheme to encode the features $x_{i}$
generated by our network from one sketch $s_{i}$ or a photo $p_{i}$.
Our hashing scheme uses a spectral normalized perception $\mathcal{F}_{SN}$
and a $\mathrm{sign}$ function to 

\begin{figure}[h]
\centering{}\includegraphics[scale=0.26]{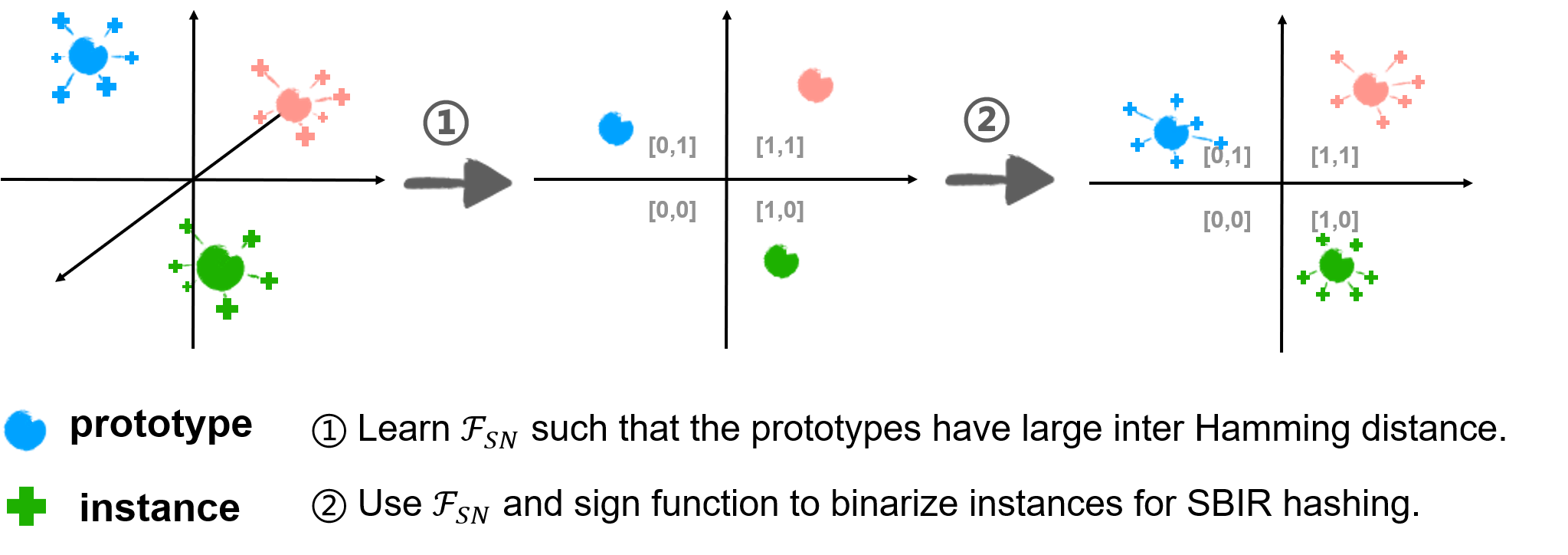}\caption{\label{fig:hash-dim-reduction}Visualization of hashing process.}
\end{figure}

\begin{table}[H]
\begin{centering}
\begin{tabular}{c|c|c}
\hline 
\textbf{\small Methods}{\small {}}  & \multicolumn{1}{>{\centering}p{2cm}|}{\textbf{\small TU-Berlin Extension}} & \multicolumn{1}{>{\centering}p{2cm}}{\textbf{\small Sketchy Extension}}\tabularnewline
\hline 
\hline 
{\small HOG\ \citep{dalal2005histograms}}  & {\small 0.091}  & {\small 0.115}\tabularnewline
{\small GF-HOG\ \citep{hu2010gradient}}  & {\small 0.119}  & {\small 0.157}\tabularnewline
{\small SHELO\ \citep{saavedra2014sketch}}  & {\small 0.123}  & {\small 0.182}\tabularnewline
{\small LKS\ \citep{saavedra2015sketch}}  & {\small 0.157}  & {\small 0.190}\tabularnewline
\hline 
{\small CCA\ \citep{via2005canonical}}  & {\small 0.366}  & {\small 0.705}\tabularnewline
{\small XQDA\ \citep{liao2015person}}  & {\small 0.201}  & {\small 0.557}\tabularnewline
{\small PLSR\ \citep{liu2018regularized}}  & {\small 0.141}  & {\small 0.462}\tabularnewline
{\small CVFL\ \citep{xie2014cross}}  & {\small 0.289 }  & {\small 0.675}\tabularnewline
\hline 
{\small Siamese CNN\ \citep{qi2016sketch}}  & {\small 0.322}  & {\small 0.481}\tabularnewline
{\small SaN\ \citep{yu2017sketch}}  & {\small 0.154}  & {\small 0.208}\tabularnewline
{\small GN Triplet\ \citep{sangkloy2016sketchy}}  & {\small 0.187}  & {\small 0.529}\tabularnewline
{\small 3D Shape\ \citep{wang2015sketch}}  & {\small 0.072}  & {\small 0.084}\tabularnewline
{\small Siamese-AlexNet\ \citep{liu2017deep}}  & {\small 0.367}  & {\small 0.518}\tabularnewline
{\small Triplet-AlexNet\ \citep{liu2017deep}}  & {\small 0.448}  & {\small 0.573}\tabularnewline
\hline 
{\small DASE+Prototypical loss\ \citep{snell2017prototypical}}  & {\small 0.433}  & {\small 0.851}\tabularnewline
{\small DASE+Triplet}\ \citep{schroff2015facenet} & {\small 0.523} & {\small 0.882}\tabularnewline
{\small DASE+Softmax}  & {\small 0.772}  & {\small 0.929}\tabularnewline
{\small DASE+A-Softmax\ \citep{liu2017sphereface}}  & {\small 0.815}  & {\small 0.938}\tabularnewline
{\small DASE+LMCL\ \citep{wang2018cosface}}  & {\small 0.837}  & {\small 0.950}\tabularnewline
\hline 
{\small ResNeXt+MEMS}  & {\small 0.798}  & {\small 0.949}\tabularnewline
\hline 
{\small DASE+MEMS (ours)}  & \textbf{\small 0.841}{\small } & \textbf{\small 0.958}\tabularnewline
\hline 
\end{tabular}
\par\end{centering}
\vspace{0.1in}
 \caption{\label{tab:Comp_SBIR}MAP results of SBIR on TU-Berlin Extension and
Sketchy Extension datasets.}
\vspace{-0.1in}
\end{table}

\noindent projects $x_{i}$ into a low-dimensional
binary space. This is a post-processing step after training CNNs. Specifically,
the $\mathcal{F}_{SN}$ is optimized by
\vspace{-0.05in}
\begin{eqnarray}
\mathcal{L}_{hash}= & \frac{1}{K(K-1)}\sum_{j}\sum_{k\ne j}\frac{\mathcal{F}_{SN}\left(\boldsymbol{c}_{j}\right)}{\left|\mathcal{F}_{SN}\left(\boldsymbol{c}_{j}\right)\right|}\cdot\frac{\mathcal{F}_{SN}\left(\boldsymbol{c}_{k}\right)}{\left|\mathcal{F}_{SN}\left(\boldsymbol{c}_{k}\right)\right|},\label{eq:hash-loss-function}
\end{eqnarray}
where $K$ is the number of categories and $|\cdot|$ represents element-wise
absolute value. Note that $\mathcal{F}_{SN}\left(\boldsymbol{x}\right)=\boldsymbol{W}_{SN}\cdot\boldsymbol{x}+\boldsymbol{b}$,
where $\boldsymbol{W}_{SN}$ is a spectral normalized matrix and $\boldsymbol{b}$
represents bias. This loss forces the prototype $c_{j}$ to have a
large Hamming distance to each other in the low-dimension space. As
$\mathcal{F}_{SN}$ is spectral normalized, the Lipschitz constant
of the mapping function from feature space to low-dimensional space
(before binarization) is 1. Thus the property of Euclidean distance
among instances, large inter-class distance and small intra-class
distance, is kept in target low-dimensional space. After $\mathrm{sign}$
function, instances in the same class will have closed binary representation.
The visualization of this process is shown in Figure\ \ref{fig:hash-dim-reduction}.

\begin{table*}[h]
\begin{centering}
\begin{tabular}{c||>{\centering}p{1.2cm}|>{\centering}p{1.2cm}|>{\centering}p{1.2cm}||>{\centering}p{1.2cm}|>{\centering}p{1.2cm}|>{\centering}p{1.2cm}}
\hline 
\multirow{2}{*}{{\textbf{\small Methods}{\small{}}}} & \multicolumn{3}{c||}{\textbf{\small TU-Berlin Extension}} & \multicolumn{3}{c}{\textbf{\small Sketchy Extension}}\tabularnewline
\cline{2-7} \cline{3-7} \cline{4-7} \cline{5-7} \cline{6-7} \cline{7-7} 
 & {\small 32 bits}  & {\small 64 bits}  & {\small 128 bits}  & {\small 32 bits}  & {\small 64 bits}  & {\small 128 bits}\tabularnewline
\hline 
{\small CMFH\ \citep{ding2014collective}}  & {\small 0.149}  & {\small 0.202}  & {\small 0.180}  & {\small 0.320}  & {\small 0.490}  & {\small 0.190}\tabularnewline
{\small CMSSH\ \citep{bronstein2010data}}  & {\small 0.121}  & {\small 0.183}  & {\small 0.175}  & {\small 0.206}  & {\small 0.211}  & {\small 0.211}\tabularnewline
{\small SCM-Seq\ \citep{zhang2014large}}  & {\small 0.211}  & {\small 0.276}  & {\small 0.332}  & {\small 0.306}  & {\small 0.417}  & {\small 0.671}\tabularnewline
{\small SCN-Orth\ \citep{zhang2014large}}  & {\small 0.217}  & {\small 0.301}  & {\small 0.263}  & {\small 0.346}  & {\small 0.536}  & {\small 0.616}\tabularnewline
{\small CVH\ \citep{kumar2011learning}}  & {\small 0.214}  & {\small 0.294}  & {\small 0.318}  & {\small 0.325}  & {\small 0.525}  & {\small 0.624}\tabularnewline
{\small SePH\ \citep{lin2015semantics}}  & {\small 0.198}  & {\small 0.270}  & {\small 0.282}  & {\small 0.534}  & {\small 0.607}  & {\small 0.640}\tabularnewline
{\small DCMH\ \citep{jiang2016deep}}  & {\small 0.274}  & {\small 0.382}  & {\small 0.425}  & {\small 0.560}  & {\small 0.622}  & {\small 0.656}\tabularnewline
{\small DSH\ \citep{liu2017deep}}  & {\small 0.358}  & {\small 0.521}  & {\small 0.570}  & {\small 0.653}  & {\small 0.711}  & {\small 0.783}\tabularnewline
{\small GDH\ \citep{zhang2018generative}}  & {\small 0.563}  & {\small 0.690}  & {\small 0.651}  & {\small 0.724}  & {\small 0.811}  & {\small 0.784}\tabularnewline
\hline 
{\small DASE+Prototypical loss\ \citep{snell2017prototypical}}  & {\small 0.381}  & {\small 0.395}  & {\small 0.400}  & {\small 0.833}  & {\small 0.845}  & {\small 0.850}\tabularnewline
{\small DASE+Triplet}\ \citep{schroff2015facenet} & {\small 0.510} & {\small 0.519} & {\small 0.527} & {\small 0.874} & {\small 0.882} & {\small 0.885}\tabularnewline
{\small DASE+Softmax}  & {\small 0.734}  & {\small 0.750}  & {\small 0.755}  & {\small 0.919}  & {\small 0.922}  & {\small 0.923}\tabularnewline
{\small DASE+A-Softmax\ \citep{liu2017sphereface}}  & {\small 0.783}  & {\small 0.800}  & {\small 0.800}  & {\small 0.925}  & {\small 0.928}  & {\small 0.927}\tabularnewline
{\small DASE+LMCL\ \citep{wang2018cosface}}  & {\small 0.762}  & {\small 0.805}  & {\small 0.818}  & {\small 0.932}  & {\small 0.935}  & {\small 0.935}\tabularnewline
\hline 
{\small DASE+MEMS (ours)}  & \textbf{\small 0.819}{\small } & \textbf{\small 0.824}{\small } & \textbf{\small 0.829}{\small } & \textbf{\small 0.948}{\small } & \textbf{\small 0.953}{\small } & \textbf{\small 0.956}\tabularnewline
\hline 
\end{tabular}
\par\end{centering}
\vspace{0.1in}
 \caption{\label{tab:Comp_cross_view}MAP results of SBIR hashing. Our model
is compared against the previous SBIR methods on TU-Berlin Extension
and Sketchy Extension. 32, 64, and 128 represents the length of generated
hashing codes.}
\end{table*}

\subsection{Results on Supervised SBIR}

\noindent \textbf{Competitors. }We compare several competitors in
Table\ \ref{tab:Comp_SBIR}: (1) hand-craft feature based models: LSK
\citep{saavedra2015sketch}, SEHLO\ \citep{saavedra2014sketch}, HOG
\citep{dalal2005histograms} and GF-HOG\ \citep{hu2010gradient}; (2)
cross-view feature embedding methods: CCA\ \citep{via2005canonical},
PLSR\ \citep{liu2018regularized}, XQDA\ \citep{liao2015person} and
CVFL\ \citep{xie2014cross}; (3) deep learning based models: 3D Shape
\citep{wang2015sketch}, Sketch-a-Net (SaN)\ \citep{yu2017sketch},
GN Triplet\ \citep{sangkloy2016sketchy} , Siamese CNN\ \citep{qi2016sketch},
Siamese-AlexNet, Triplet-AlexNet\ \citep{liu2017deep}. (4) Prototypical
loss\ \citep{snell2017prototypical}, Triplet loss\ \citep{schroff2015facenet}, Softmax, A-Softmax\ \citep{liu2017sphereface} and LMCL\ \citep{wang2018cosface} implemented with the same backbone network as our method. We use hyperparameters of A-Softmax\ \citep{liu2017sphereface} and LMCL\ \citep{wang2018cosface} proposed in their papers. The Mean Average Precision (MAP) is reported.

\noindent \textbf{Results. }The results are summarized in Table~\ref{tab:Comp_SBIR}.
Obviously, our model outperforms all the competitors by a very large
margin. It achieves a MAP improvement of 0.393/0.385 over the state-of-the-art
real-valued based method -- Triplet-AlexNet. This demonstrates the
efficacy of our model. Note that the improved performance is due to
the novel structure, and the MEMS loss function used here. We give
further analysis in the ablation study in Section \ref{sec:ablation}.

The effect of each component in our framework can be found in Table
\ref{tab:Comp_SBIR}. We can conclude that the proposed MEMS loss
and DASE module both improve the performance on SBIR task significantly.

\noindent {\small } 
\begin{table}
\begin{centering}
\begin{tabular}{c|c|>{\centering}p{1.3cm}|>{\centering}p{1.3cm}|c}
\hline 
\textbf{\small Backbone}{\small } & \textbf{\small Methods}{\small } & \textbf{\small TU-Berlin Extension}{\small } & \textbf{\small Sketchy Extension}{\small } & \textbf{\small FLOPS}\tabularnewline
\hline 
\hline 
\multirow{2}{*}{{\small AlexNet}} & {\small DSH\ \citep{liu2017deep}}  & {\small 0.570}  & {\small 0.783}  & {\small 1.65G}\tabularnewline
 & {\small DASE+MEMS (ours)}  & {\small 0.586}  & {\small 0.875}  & {\small 1.13G}\tabularnewline
\hline 
\multirow{2}{*}{{\small Resnet-18}} & {\small GDH\ \citep{zhang2018generative}}  & {\small 0.690}  & {\small 0.811}  & {\small 16.93G}\tabularnewline
 & {\small DASE+MEMS (ours)}  & {\small 0.736}  & {\small 0.940}  & {\small 1.83G}\tabularnewline
\hline 
\end{tabular}
\par\end{centering}
\vspace{0.1in}

{\small \caption{\label{tab:compare_same_backbone}MAP results achieved with same backbone
network. DSH\ \citep{liu2017deep} modifies AlexNet to a semi-heterogeneous
network; GAN is used in GDH\ \citep{zhang2018generative}besides backbone
network; we only append backbone network with DASE modules.}
}{\small\par}
\end{table}

\subsection{Results on Supervised SBIR Hashing}

\noindent \textbf{Competitors. }(1) Our hashing model is compared
against 8 cross-modal hashing methods: Collective Matrix Factorization
Hashing (CMFH)\ \citep{ding2014collective}, Cross-Modal Semi-Supervised
Hashing (CMSSH)\ \citep{bronstein2010data}, Cross-View Hashing(CVH)
\citep{kumar2011learning}, Semantic Correlation Maximization (SCMSeq
and SCM-Orth)\ \citep{zhang2014large}, Semantics-Preserving Hashing(SePH)
\citep{lin2015semantics}, Deep CrossModality Hashing (DCMH)\ \citep{jiang2016deep},
Deep Sketch Hash (DSH)\ \citep{liu2017deep} and Generative Domain-Migration
Hashing (GDH)\ \citep{zhang2018generative}. (2) Other softmax based
loss functions implemented with the same backbone as our method: Prototypical
loss\ \citep{snell2017prototypical}, Triplet loss\ \citep{schroff2015facenet}, Softmax, A-Softmax\ \citep{liu2017sphereface}
and LMCL\ \citep{wang2018cosface}. We use hyperparameters of A-Softmax
\citep{liu2017sphereface} and LMCL\ \citep{wang2018cosface} proposed
in their papers. We still report the MAP.

\noindent \textbf{Training cost.} Our hashing scheme is taken as a
post-processing step, in order to make our framework comparable to
previous hashing based SBIR models. With our computed features, the
hashing scheme is trained for 10000 steps; and the whole process can be
finished in 1 minute on our computer.

\begin{figure*}
\centering{}%
\begin{tabular}{c|c}
\hline 
\textbf{\small TUBerlin Extension}{\small } & \textbf{\small Sketchy Extension}\tabularnewline
\hline 
\includegraphics[width=8cm]{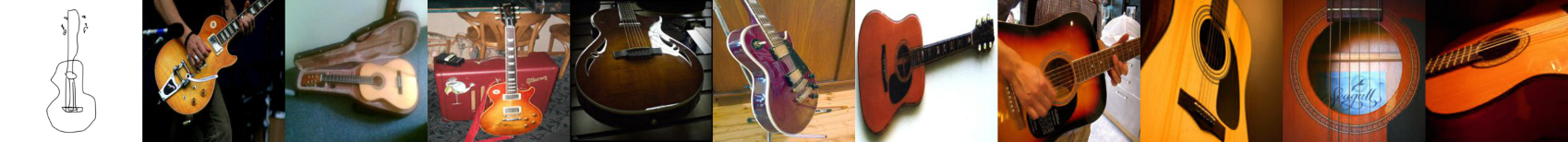}  & \includegraphics[width=8cm]{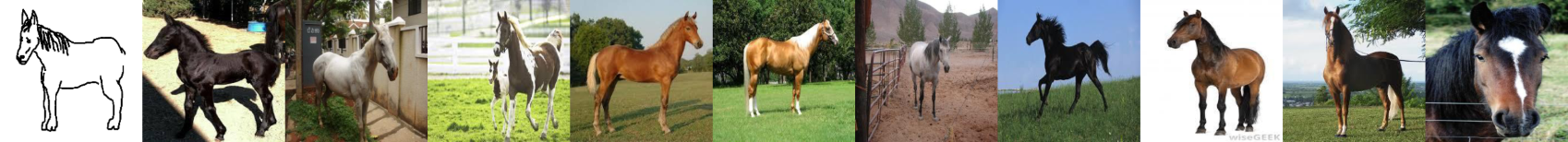}\tabularnewline
\includegraphics[width=8cm]{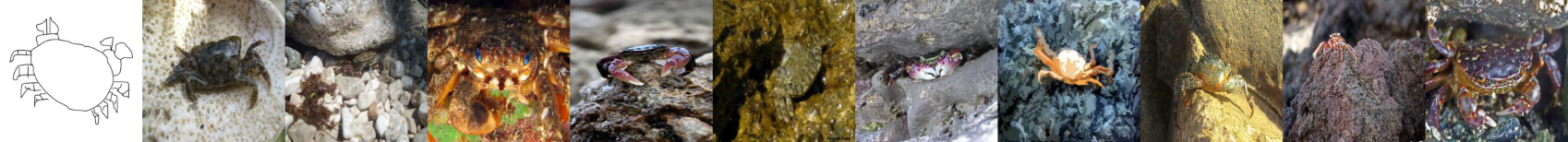}  & \includegraphics[width=8cm]{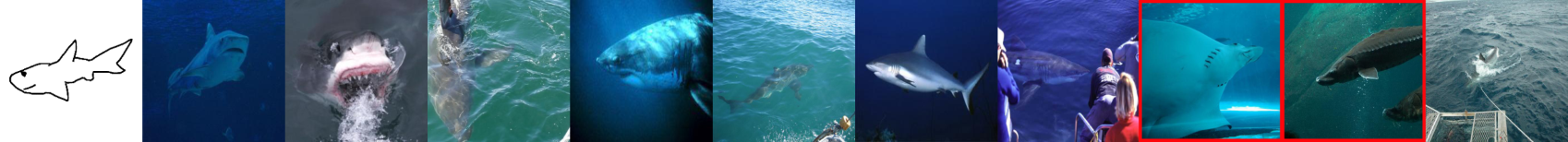}\tabularnewline
\includegraphics[width=8cm]{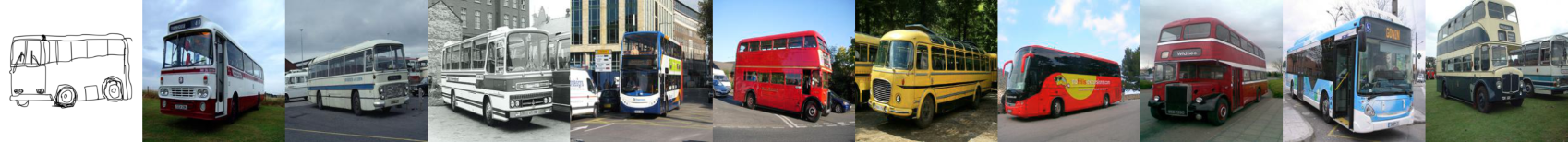}  & \includegraphics[width=8cm]{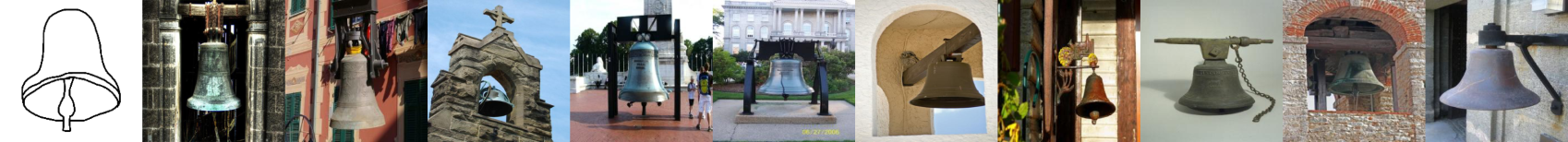}\tabularnewline
\includegraphics[width=8cm]{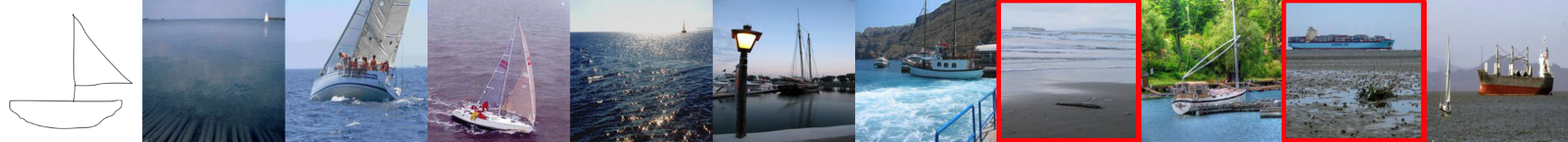}  & \includegraphics[width=8cm]{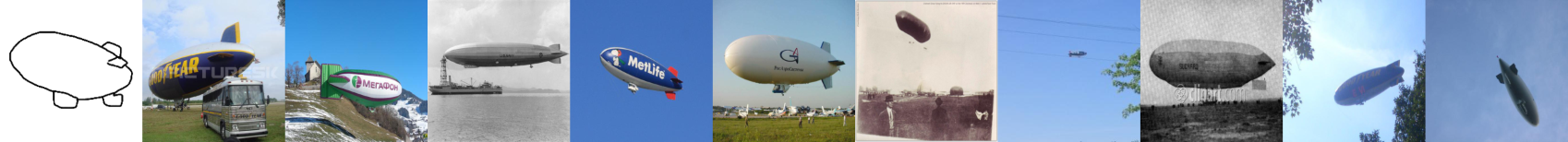}\tabularnewline
\includegraphics[width=8cm]{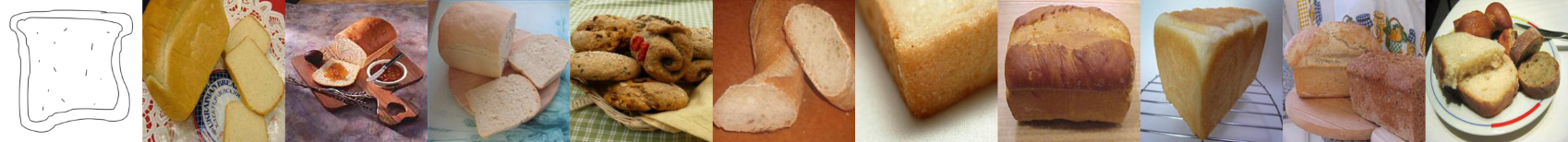}  & \includegraphics[width=8cm]{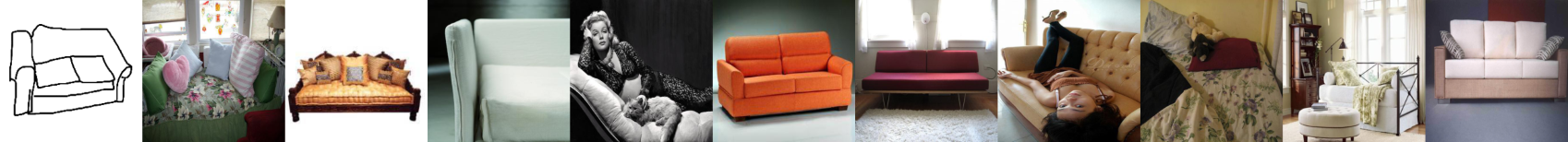}\tabularnewline
\hline 
\end{tabular}\caption{\label{tab:Visulization-of-retrived}Visulization of retrieved images
on both datasets. The red panes represent false retrieval results.}
\end{figure*}

\noindent \textbf{Results. }We summarize our results in Table\ \ref{tab:Comp_cross_view}.
Our method achieves the best performance among all hashing-based methods
and cross-modal learning methods. Critically, our model improves MAP
with a scale over 0.13 in all conditions compared with GDH\ \citep{zhang2018generative}
which is the state-of-the-art method on this task. This further demonstrates
the effectiveness of our framework in the SBIR hashing task. Some query examples with top-10 retrieval results are shown in Figure\ \ref{tab:Visulization-of-retrived}.

We also highlight that, although angular margin losses, particularly
LMCL, can achieve comparable performance with MEMS in SBIR tasks,
they suffer from a significant degradation in SBIR hashing. This degradation
might be due to the property of angular distance, as it is more difficult
to maintain in the dimension reduction mapping than Euclidean distance.

\noindent \textbf{Comparison using same backbone.} To emphasize the
efficiency and efficacy of our model, we compare the MAP with state-of-the-art
methods DSH\ \citep{liu2017deep} and GDH\ \citep{zhang2018generative}
using the same backbone networks, which is Alexnet and Resnet-18 respectively.
The result is displayed in Table. \ref{tab:compare_same_backbone}.
The proposed method achieves higher MAP than other methods with much
lower computation consumption.{\small  }{\small\par}

Besides, our model can be trained efficiently. We only need to train
a single network in an end-to-end manner without any domain translation
process. On the contrary, GDH\ \citep{zhang2018generative} utilized
the cycle-consistent GANs to transfer sketches into photos. DSH\ \citep{liu2017deep}
requires the pre-computed edge maps to bridge the gap between sketches
and photos, and semantic representation (wordvec) is used as prior
of inter-relationship among categories.

\subsection{\label{subsec:Ablation-Study}Ablation Study\textcolor{red}{{} }}

\label{sec:ablation} To further investigate the propose approach,
we conduct a series of ablation studies on both datasets.

\begin{figure}
\begin{centering}
\includegraphics[scale=0.45]{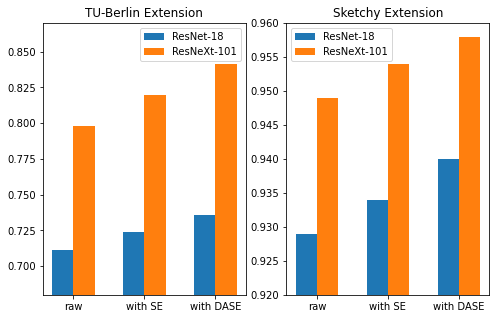} 
\par\end{centering}
\caption{\label{tab:expr_csemodule}Performance of different network variants;
all of which are pre-trained on ImageNet. We use MEMS loss with $m=4$
to train these networks.}
\vspace{-0.15in}
\end{figure}

\noindent \textbf{Network modules}. We compare two types of CNNs as
well as their variants with SE modules\ \citep{hu2017squeeze}
and the newly DASE modules. Intrinsically, the SE and DASE module can
enhance the ability to learn different attention over feature channels,
and thus enable a dynamic and implicit feature selection mechanism
to our networks. The MAP results are shown in Figure~\ref{tab:expr_csemodule}.
Both networks are trained and tested in the same setting; and \emph{the
same MEMS loss is used for all the networks.} We can find that SE
module enhances the ability of CNNs to process inputs from multi-domains.
Moreover, our DASE module is better than SE module on SBIR task, since
the auxiliary binary code is introduced to make SE better learn to
select important sketch/photo feature channels. DASE module can improve
the performance of both deep and shallow networks. Particularly, on Resnet-18/ResNeXt-101, the auxiliary DASE modules have brought the MAP increase with 0.025/0.043 on TU-Berlin Extension and 0.011/0.009 on Sketchy Extension, respectively.

\begin{figure}
\hspace{-0.5cm}\includegraphics[width=8.6cm]{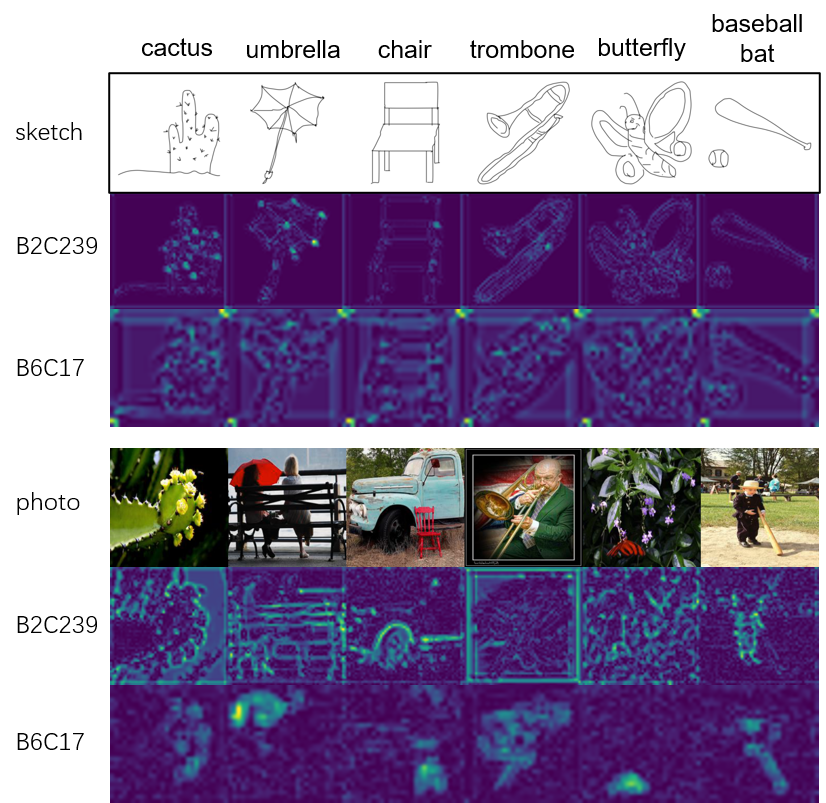}

\caption{\label{fig:Visualization-of-feature}Visualization of feature maps
amplified/suppressed by DASE. B2C239 refers to the 239-th channel
of input feature map to the DASE module in second convolution block.
B6C17 refers to the 17-th channel and 6-th convolution block.}
\end{figure}

We also visualize the input feature maps of DASE modules in Figure~\ref{fig:Visualization-of-feature}.
It is noticable that some convolution kernels that can effectively
extract content-related features from sketches might fail to extract
such features from photos, and vice versa. Specifically, features
in B2C239 generally respond to dark areas, and thus miss the key object
in photos but capture crucial corners in sketches. On the contrary,
in B6C17, sketch features show severe corner and edge effects which
distort the semantic, while photo features respond to correct area.
With prior knowledge, the DASE module can learn to amplify content-related
features and suppress style-related features. The weight in the second
fully-connected layer in DASE module associated with the binary code
about input domain and B2C239 is -0.32, smaller than most channels,
indicating that this channel is relatively amplified/suppressed when the input
is a sketch/photo, respectively. As for B6C17, this weight is 0.30,
and is larger than most channels, meaning that the photo features
are amplified and the sketch features are suppressed.

\begin{figure}
\centering{}\includegraphics[scale=0.6]{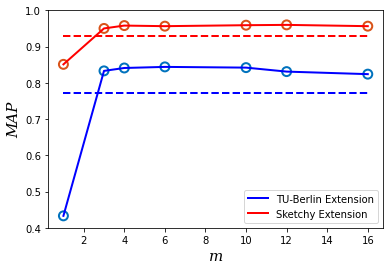}
\vspace{-0.15in}
\caption{\label{fig:ablation_diff_m}The MAP on two datasets with varying $m$
in MEMS loss. The dashed lines represent the results of softmax on
each dataset.}
\end{figure}

\noindent \textbf{The effects of different margin $m$.} Performance
of our model grows with the increase of value $m$, because it forces
the network to learn discriminative representations. But when $m$
is too large, \emph{e.g.,} $m>4$, the performance stops rising and
even begins falling, which is shown in Figure\ \ref{fig:ablation_diff_m}.

\begin{figure}
\centering{}%
\begin{tabular}{ccc}
\hspace{-0.15in}\includegraphics[scale=0.27]{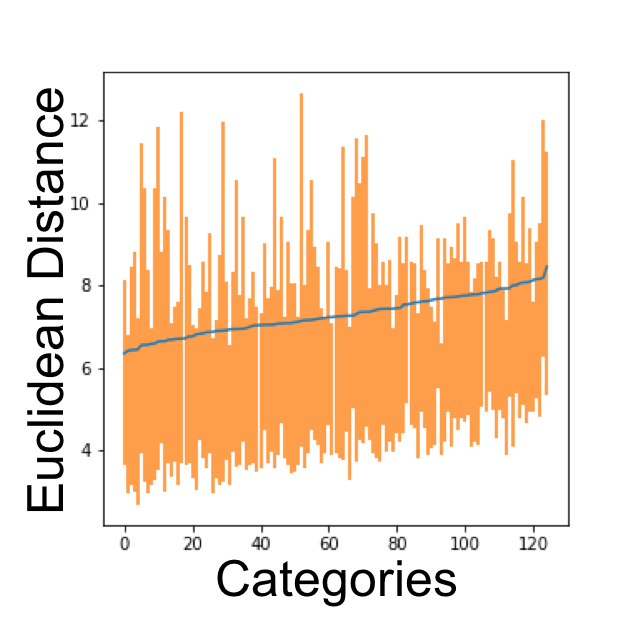}  & \hspace{-0.15in}\includegraphics[scale=0.27]{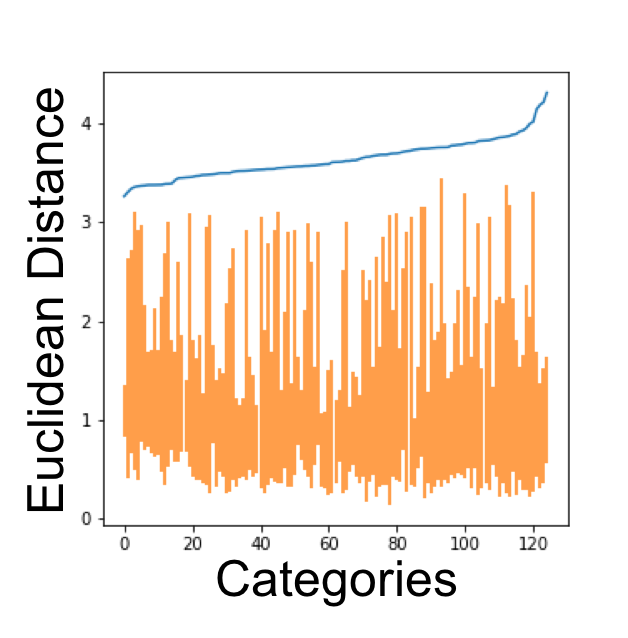}  & \hspace{-0.15in}\includegraphics[scale=0.27]{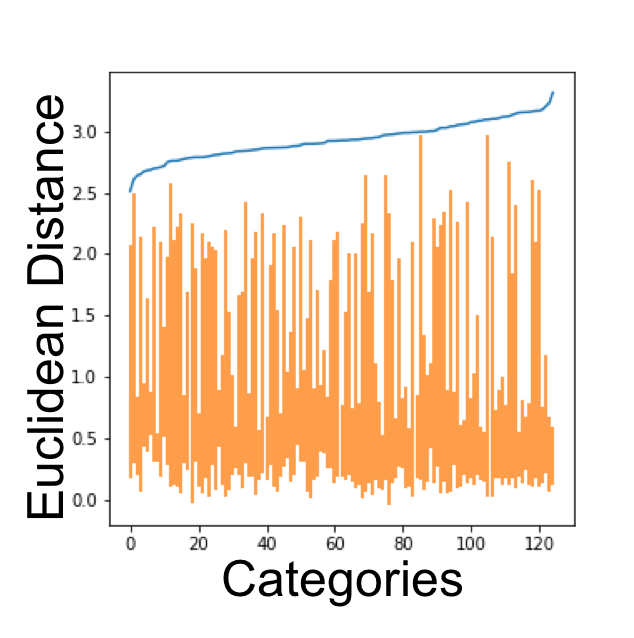}\tabularnewline
\hspace{-0.15in}(a) $m=1$  & \hspace{-0.15in}(b) $m=4$  & \hspace{-0.15in}(d) $m=16$\tabularnewline
\end{tabular}\vspace{0.1in}
 \caption{\label{fig:diff_m_sketchy} Visualization of inter-class and intra-class distance on Sketchy Extension. The blue lines denote average minimum inter-class distance. The orange bars represent the distribution of intra-class distance of each category.}
\end{figure}

\noindent As revealed in Figure\ \ref{fig:diff_m_sketchy}: (1) when
$m=1$, the intra-class distances of part of instances are greater
than the minimum inter-class distance, which leads to bad retrieval
performance; (2) when $m=4$ and $m=16$, the minimum inter-class
distances are greater than intra-class distances generally, which
explains the better performance in these cases. Nevertheless, the
standard deviation within some categories does not decrease with larger
$m$ due to intrinsic variance, explaining why the performance stops
rising or begins falling when $m$ is overly large.

\begin{figure}
\centering{}\includegraphics[scale=0.45]{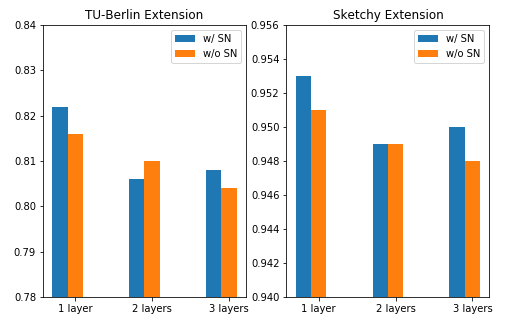}
\caption{\label{fig:ablation_hashing}Comparison of MAP results with different settings of $\mathcal{F}_{SN}$.}
\end{figure}

\noindent \textbf{Analysis of the mapping in hashing}. We train a spectral normalized fully-connected layer $\mathcal{F}_{SN}$ for hashing. But the mapping can be modeled by other structures. We try fully-connected networks with different depth and settings (w./w.o. spectral normalization) for the hashing task. 
Note that if we adopt multi-layer networks, the dimension of intermediate layers is set to 256, and the activation function is Rectified Linear Unit, as the Lipschitz constant of ReLU is smaller than 1. The results of MAP on both datasets are shown in Figure\ \ref{fig:ablation_hashing}. We can found that multi-layer networks have worse performance than one-layer linear mapping. Moreover, the spectral normalized networks work better generally, which shows their advantage in maintaining discrimination property.

\section{Conclusion}

In this paper, we have introduced two innovations, including novel
network architecture and a loss function. The Domain-Aware Squeeze-and-Excitation
(DASE) network allows us to incorporate the domain information of
each sample explicitly, while the proposed Multiplicative Euclidean
Margin Softmax (MEMS) Loss enables us to learn highly discriminative
features which facilitate highly accurate SBIR. Both the architecture
and loss are intuitive and simple to implement. On two popular benchmark
SBIR datasets, the proposed model has achieved new state-of-the-art
results. 

\begin{acks}
This work was supported by the Natural Science Foundation of Guangdong Province(No.2020A1515010711) and the Special Foundation for the Development of Strategic Emerging Industries of Shenzhen(Nos. JCYJ20200109143010272 and JCYJ20200109143035495).
This work was supported in part by Science and Technology Commission of Shanghai Municipality Projects (19511120700, 2021SHZDZX0103).
\end{acks}

\bibliographystyle{ACM-Reference-Format}
\bibliography{egbib}

\clearpage{}

\onecolumn

\section{Appendix: Theoretical Analysis of MEMS Loss}

In this section, we will (1) give a formal definition of maximum intra-class
distance and minimum inter-class distance; (2) show that for margin
$m$ in MEMS loss, $m\ge2+\sqrt{3}$ is sufficient and necessary to
ensure that the maximum intra-class distance is smaller than minimum
inter-class distance, regardless of the number of categories.

\subsection{Definition}

Since we treat both sketch and photo as instances, we define the merged
dataset as:

\begin{equation}
\mathcal{D}=\left\{ \left(\boldsymbol{x}_{i},y_{\boldsymbol{x}_{i}}\right)\vert y_{\boldsymbol{x}_{i}}\in\mathcal{Y}\right\} {}_{i=1}^{n_{1}+n_{2}}\quad\text{where}\quad\begin{cases}
\boldsymbol{x}_{i}=\mathcal{F}_{p}\left(p_{i}\right),y_{\boldsymbol{x}_{i}}=y_{p_{i}} & \text{if}\quad i\le n_{1}\\
\boldsymbol{x}_{i}=\mathcal{F}_{s}\left(s_{i-n_{1}}\right),y_{\boldsymbol{x}_{i}}=y_{s_{i-n_{1}}} & \text{if}\quad i>n_{1}
\end{cases}
\end{equation}
where $\mathcal{F}_{p}$ and $\mathcal{F}_{s}$ are mappings that
map photos/sketches into a feature space; $n_1,n_2$ represent the number of photos and sketches; $p,s$ and $y$ represent photo, sketch and category respectively. They are detailedly illustrated in Sec. 3. For convenience, we also define $\mathcal{D}_{y}=\{x|(x,y_{x})\in\mathcal{D}\wedge y_{x}=y\}$.

\paragraph{Maximum Intra-class Distance and Minimum Inter-class Distance}

For category $y\in\mathcal{Y}$, the maximum intra-class distance
can be defined by 
\[
\max_{\boldsymbol{x},\boldsymbol{x}'\in\mathcal{D}_{y}}\mathrm{d}\left(\boldsymbol{x},\boldsymbol{x}'\right)
\]
and the minimum inter-class distance: 
\[
\min_{\boldsymbol{x}\in\mathcal{D}_{y},\boldsymbol{x}'\in\left(\bigcup_{y'\ne y}\mathcal{D}_{y'}\right)}\mathrm{d}\left(\boldsymbol{x},\boldsymbol{x}'\right)
\]

Here we give a formulation of our objective, which is the maximum
intra-class distance being smaller than minimum inter-class distance,
by proposition $P_{1}$:

\begin{equation}
\forall y\in\mathcal{Y},\quad\max_{\boldsymbol{x},\boldsymbol{x}'\in\mathcal{D}_{y}}\mathrm{d}\left(\boldsymbol{x},\boldsymbol{x}'\right)\le\min_{\boldsymbol{x}\in\mathcal{D}_{y},\boldsymbol{x}'\in\left(\bigcup_{y'\ne y}\mathcal{D}_{y'}\right)}\mathrm{d}\left(\boldsymbol{x},\boldsymbol{x}'\right)\label{eq:target-appendix}
\end{equation}

\paragraph{Solve Problem with MEMS }

Instead of optimizing the distance among instances directly as indicated
by eq. \ref{eq:target-appendix}, the proposed MEMS loss uses prototypes
$\left\{ \boldsymbol{c}_{y}\right\} _{y\in\mathcal{Y}}\subset\mathcal{X}$
to characterize the distribution of instances in feature space. If
this MEMS loss is well optimized, instances will be closer to their
corresponding prototype than other prototypes in feature space. This
relationship can be described as 
\begin{equation}
\forall\left(\boldsymbol{x},y_{\boldsymbol{x}}\right)\in\mathcal{D},\forall y\in\mathcal{Y}\wedge y\ne y_{\boldsymbol{x}},\quad m\cdot\mathrm{d}\left(\boldsymbol{x},\boldsymbol{c}_{y_{\boldsymbol{x}}}\right)\le\mathrm{d}\left(\boldsymbol{x},\boldsymbol{c}_{y}\right)\label{eq:multiplicative-euclidean-distance-appendix}
\end{equation}

For convenience, we denote $\mathcal{R}_{y,y'}$ as a region where
\[
\boldsymbol{x}\in\mathcal{R}_{y,y'}\quad\Leftrightarrow\quad m\cdot\mathrm{d}\left(\boldsymbol{x},\boldsymbol{c}_{y}\right)\le\mathrm{d}\left(\boldsymbol{x},\boldsymbol{c}_{y'}\right)
\]
Also, we denote $\mathcal{R}_{y}$ as a region where 
\[
\boldsymbol{x}\in\mathcal{R}_{y}\quad\Leftrightarrow\quad\forall y'\in\mathcal{Y}\wedge y'\ne y,m\cdot\mathrm{d}\left(\boldsymbol{x},\boldsymbol{c}_{y}\right)\le\mathrm{d}\left(\boldsymbol{x},\boldsymbol{c}_{y'}\right)
\]
It is easy to prove that $\mathcal{R}_{y}=\bigcap_{y'\ne y}\mathcal{R}_{y,y'}$.
Note that if eq. \ref{eq:multiplicative-euclidean-distance-appendix}
holds, 
\[
\forall\left(\boldsymbol{x},y_{\boldsymbol{x}}\right)\in\mathcal{D},\quad\boldsymbol{x}\in\mathcal{R}_{y_{\boldsymbol{x}}}
\]
and thus we can derive a sufficient condition for $P_{1}$ (eq. \ref{eq:target-appendix}):
\begin{equation}
\forall y\in\mathcal{Y},\quad\max_{\boldsymbol{x},\boldsymbol{x}'\in\mathcal{R}_{y}}\mathrm{d}\left(\boldsymbol{x},\boldsymbol{x}'\right)\le\min_{\boldsymbol{x}\in\mathcal{R}_{y},\boldsymbol{x}'\in\left(\bigcup_{y'\ne y}\mathcal{R}_{y'}\right)}\mathrm{d}\left(\boldsymbol{x},\boldsymbol{x}'\right)\label{eq:target-sufficient-appendix}
\end{equation}
We denote eq. \ref{eq:target-sufficient-appendix} as proposition
$P_{2}$.

\subsection{Finding Boundaries for $m$}

The later induction is based on the assumption that our MEMS loss
can be well optimized, \textit{i.e.} eq. \ref{eq:multiplicative-euclidean-distance-appendix}
holds , and the assumption that $m\ge1$. Now the question is: what
is the range of $m$ that is sufficient and necessary for $P_{2}$?
In the rest of this section, we firstly calculate the closed form
of $\mathcal{R}_{y,y'}$ and then prove that if $m=m_{0}\Rightarrow P_{2}$,
then $m=m_{0}+\varepsilon\Rightarrow P_{2},\forall\varepsilon>0$.
To this end, we only need to find the lower bound of $m$, with regard
to the number of categories: $m_{\min}^{\left(\left|\mathcal{Y}\right|\right)}$.
Next, we prove $m_{\min}^{\left(2\right)}=2+\sqrt{3}$. Finally, we
prove $m_{\min}^{\left(k\right)}=2+\sqrt{3},\forall k\ge3$ is sufficient
and necessary for $P_{2}$.

\paragraph{Lemma 1}

If $\mathrm{d}(\boldsymbol{x},\boldsymbol{x}')=\left\Vert \boldsymbol{x}-\boldsymbol{x}'\right\Vert _{2}$,
$\mathcal{R}_{y,y'}$ is a n-ball (ball in n-dimensional space) with
center $\left(\boldsymbol{c}_{y}+\frac{\boldsymbol{c}_{y}-\boldsymbol{c}_{y'}}{m^{2}-1}\right)$
and radius $\left(\frac{m}{m^{2}-1}\right)\left\Vert \boldsymbol{c}_{y}-\boldsymbol{c}_{y'}\right\Vert _{2}$

\subparagraph{Proof}

If $\boldsymbol{x}\in\mathcal{R}_{y,y'}$,

\begin{eqnarray*}
 & m\left\Vert \boldsymbol{x}-\boldsymbol{c}_{y}\right\Vert _{2}\le\left\Vert \boldsymbol{x}-\boldsymbol{c}_{y'}\right\Vert _{2}\\
\Leftrightarrow & m^{2}\left\Vert \boldsymbol{x}-\boldsymbol{c}_{y}\right\Vert _{2}\le\left\Vert \boldsymbol{x}-\boldsymbol{c}_{y'}\right\Vert _{2}\\
\Leftrightarrow & m^{2}\left(\boldsymbol{x}^{T}\boldsymbol{x}-2\boldsymbol{c}_{y}^{T}\boldsymbol{x}+\boldsymbol{c}_{y}^{T}\boldsymbol{c}_{y}\right)\le\boldsymbol{x}^{T}\boldsymbol{x}-2\boldsymbol{c}_{y'}^{T}\boldsymbol{x}+\boldsymbol{c}_{y'}^{T}\boldsymbol{c}_{y'}\\
\Leftrightarrow & \left(m^{2}-1\right)\boldsymbol{x}^{T}\boldsymbol{x}-2\left(m^{2}\boldsymbol{c}_{y}^{T}-\boldsymbol{c}_{y}^{T}\right)\boldsymbol{x}\le\boldsymbol{c}_{y'}^{T}\boldsymbol{c}_{y'}-m^{2}\boldsymbol{c}_{y}^{T}\boldsymbol{c}_{y}\\
\Leftrightarrow & \left\Vert \boldsymbol{x}-\frac{m^{2}\boldsymbol{c}_{y}-\boldsymbol{c}_{y'}}{m^{2}-1}\right\Vert _{2}^{2}\le\frac{\left(m^{2}-1\right)\left(\boldsymbol{c}_{y'}^{T}\boldsymbol{c}_{y'}-m^{2}\boldsymbol{c}_{y}^{T}\boldsymbol{c}_{y}\right)+\left\Vert m^{2}\boldsymbol{c}_{y}-\boldsymbol{c}_{y'}\right\Vert _{2}^{2}}{\left(m^{2}-1\right)^{2}}\\
\Leftrightarrow & \left\Vert \boldsymbol{x}-\frac{\left(m^{2}-1\right)\boldsymbol{c}_{y}+\boldsymbol{c}_{y}-\boldsymbol{c}_{y'}}{m^{2}-1}\right\Vert _{2}^{2}\le\left(\frac{m}{m^{2}-1}\right)^{2}\left\Vert \boldsymbol{c}_{y}-\boldsymbol{c}_{y'}\right\Vert _{2}^{2}\\
\Leftrightarrow & \left\Vert \boldsymbol{x}-\left(\boldsymbol{c}_{y}+\frac{\boldsymbol{c}_{y}-\boldsymbol{c}_{y'}}{m^{2}-1}\right)\right\Vert _{2}\le\left(\frac{m}{m^{2}-1}\right)\left\Vert \boldsymbol{c}_{y}-\boldsymbol{c}_{y'}\right\Vert _{2}
\end{eqnarray*}

\paragraph{Lemma 2}

If $m=m_{0}\Rightarrow P_{2}$, then $m=m_{0}+\varepsilon\Rightarrow P_{2},\forall\varepsilon>0$

\subparagraph{Proof}

With $m=m_{0}$ slightly expanding to $m=m_{0}+\varepsilon$, region
$\mathcal{R}_{y,y'}$ becomes $\mathcal{R}'_{y,y'}$, where 
\[
\forall\boldsymbol{x}\in\mathcal{R}'_{y,y'},\quad\mathrm{d}\left(\boldsymbol{x},\boldsymbol{c}_{y'}\right)\ge\left(m_{0}+\varepsilon\right)\mathrm{d}\left(\boldsymbol{x},\boldsymbol{c}_{y}\right)\ge m_{0}\mathrm{d}\left(\boldsymbol{x},\boldsymbol{c}_{y}\right)
\]
So we can conclude that $\mathcal{R}'_{y,y'}\subseteq\mathcal{R}_{y,y'}$.
Thus 
\begin{equation}
\forall y\in\mathcal{Y},\quad\mathcal{R}'_{y}=\bigcap_{y'\ne y}\mathcal{R}'_{y,y'}\subseteq\bigcap_{y'\ne y}\mathcal{R}{}_{y,y'}=\mathcal{R}_{y}\label{eq:ineq_rr}
\end{equation}

Now we rewrite $P_{2}$ as 
\begin{equation}
\forall y\in\mathcal{Y},\forall y'\in\mathcal{Y}\wedge y'\ne y,\quad\max_{\boldsymbol{x},\boldsymbol{x}'\in\mathcal{R}_{y}}\mathrm{d}\left(\boldsymbol{x},\boldsymbol{x}'\right)\le\min_{\boldsymbol{x}\in\mathcal{R}_{y},\boldsymbol{x}'\in\mathcal{R}_{y'}}\mathrm{d}\left(\boldsymbol{x},\boldsymbol{x}'\right)\label{eq:target-sufficient-2-appendix}
\end{equation}

Since eq. \ref{eq:ineq_rr}, 
\begin{eqnarray*}
\max_{\boldsymbol{x},\boldsymbol{x}'\in\mathcal{R}'_{y}}\mathrm{d}\left(\boldsymbol{x},\boldsymbol{x}'\right) & \le & \max_{\boldsymbol{x},\boldsymbol{x}'\in\mathcal{R}_{y}}\mathrm{d}\left(\boldsymbol{x},\boldsymbol{x}'\right)\\
\min_{\boldsymbol{x}\in\mathcal{R}'_{y},\boldsymbol{x}'\in\mathcal{R}'_{y'}}\mathrm{d}\left(\boldsymbol{x},\boldsymbol{x}'\right) & \ge & \min_{\boldsymbol{x}\in\mathcal{R}_{y},\boldsymbol{x}'\in\mathcal{R}_{y'}}\mathrm{d}\left(\boldsymbol{x},\boldsymbol{x}'\right)
\end{eqnarray*}
which means if eq. \ref{eq:target-sufficient-2-appendix} holds for
$m=m_{0}$, it also holds for $m=m_{0}+\varepsilon$.

\paragraph{Lemma 3}

$m_{\min}^{\left(2\right)}=2+\sqrt{3}$ is sufficient and necessary
for $P_{2}$.

\subparagraph{Proof}

We can write $\mathcal{Y}=\left\{ 1,2\right\} $. Now $\mathcal{R}_{1}=\mathcal{R}_{1,2},\mathcal{R}_{2}=\mathcal{R}_{2,1}$,
which are two n-balls with same radius and different centers. The
maximum intra-class distance is the diameter of each n-ball: 
\[
\forall y\in\mathcal{Y},\quad\max_{\boldsymbol{x},\boldsymbol{x}'\in\mathcal{R}_{y}}\mathrm{d}\left(\boldsymbol{x},\boldsymbol{x}'\right)=\left(\frac{2m}{m^{2}-1}\right)\left\Vert \boldsymbol{c}_{1}-\boldsymbol{c}_{2}\right\Vert _{2}
\]

The minimum inter-class distance is the distance between two centers
minus the diameter: 
\begin{eqnarray*}
\forall y\in\mathcal{Y}, & \min_{\boldsymbol{x}\in\mathcal{R}_{y},\boldsymbol{x}'\in\left(\bigcup_{y'\ne y}\mathcal{R}_{y'}\right)}\mathrm{d}\left(\boldsymbol{x},\boldsymbol{x}'\right) & =\left\Vert \left(\boldsymbol{c}_{1}+\frac{\boldsymbol{c}_{1}-\boldsymbol{c}_{2}}{m^{2}-1}\right)-\left(\boldsymbol{c}_{2}+\frac{\boldsymbol{c}_{2}-\boldsymbol{c}_{1}}{m^{2}-1}\right)\right\Vert _{2}-\left(\frac{2m}{m^{2}-1}\right)\left\Vert \boldsymbol{c}_{1}-\boldsymbol{c}_{2}\right\Vert _{2}\\
 &  & =\left(\frac{m^{2}-2m+1}{m^{2}-1}\right)\left\Vert \boldsymbol{c}_{1}-\boldsymbol{c}_{2}\right\Vert _{2}\\
\end{eqnarray*}

Let $\left(\frac{m^{2}-2m+1}{m^{2}-1}\right)\left\Vert \boldsymbol{c}_{1}-\boldsymbol{c}_{2}\right\Vert _{2}\ge\left(\frac{2m}{m^{2}-1}\right)\left\Vert \boldsymbol{c}_{1}-\boldsymbol{c}_{2}\right\Vert _{2}$,
we can get the result $m\ge2+\sqrt{3}$ or $m\le2-\sqrt{3}$. We abandon
the latter solution since only when $m\ge1$ does it make sense. So
in binary class case, $m\ge m_{\min}^{\left(2\right)}=2+\sqrt{3}$
if sufficient and necessary for $P_{2}$.

\paragraph{Lemma 4}

$m_{\min}^{\left(k\right)}=2+\sqrt{3},\forall k\ge3$ is necessary
for $P_{2}$.

\subparagraph{Proof}

Consider an extreme condition, where two prototypes $\boldsymbol{c}_{y_{a}},\boldsymbol{c}_{y_{b}}$
are far from the other prototypes. We notice that 
\[
\forall y\in\mathcal{Y}\wedge y\ne y_{a}\wedge y\ne y_{b},\left\Vert \boldsymbol{c}_{y_{a}}-\boldsymbol{c}_{y}\right\Vert _{2}\ge\frac{m+1}{m-1}\left\Vert \boldsymbol{c}_{y_{a}}-\boldsymbol{c}_{y_{b}}\right\Vert _{2}\Rightarrow\min_{\boldsymbol{x}\in\mathcal{R}_{y_{a},y}}\mathrm{d}\left(\boldsymbol{x},\boldsymbol{c}_{y_{a}}\right)\ge\max_{\boldsymbol{x}\in\mathcal{R}_{y_{a},y_{b}}}\mathrm{d}\left(\boldsymbol{x},\boldsymbol{c}_{y_{a}}\right)\Rightarrow\mathcal{R}_{y_{a},y_{b}}\subseteq\mathcal{R}_{y_{a},y}
\]
Since we have no constraints on location of prototypes, this condition
can always be likely to hold, regardless of the value of $m$. When
all the rest prototypes satisfy this condition for both $\boldsymbol{c}_{y_{a}},\boldsymbol{c}_{y_{b}}$,
we have $\mathcal{R}_{y_{a}}=\mathcal{R}_{y_{a},y_{b}}$ and $\mathcal{R}_{y_{b}}=\mathcal{R}_{y_{b},y_{a}}$
, which is same as in binary case. So $m\ge m_{\min}^{\left(2\right)}$
becomes necessary to ensure the correctness of $P_{2}$, and thus
Lemma 4 is true.

\paragraph{Lemma 5}

$m_{\min}^{\left(k\right)}=2+\sqrt{3},\forall k\ge3$ is sufficient
for $P_{2}$.

\subparagraph{Proof}

If we want to prove that $P_{2}$ (eq. \ref{eq:target-sufficient-2-appendix}) holds, we have to show that every distinct pair $\left(y_{a},y_{b}\right)$ satisfy
$\max_{\boldsymbol{x},\boldsymbol{x}'\in\mathcal{R}_{y_{a}}}\mathrm{d}\left(\boldsymbol{x},\boldsymbol{x}'\right)\le\min_{x\in\mathcal{R}_{y_{a}},x'\in\mathcal{R}_{y_{b}}}\mathrm{d}\left(\boldsymbol{x},\boldsymbol{x}'\right)$.
We remove a category $y_{c}$, where $y_{c}\ne y_{a}$ and $y_{c}\ne y_{b}$,
from $\mathcal{Y}$ and forms $\mathcal{Y}'$ such that $\left|\mathcal{Y}'\right|=\left|\mathcal{Y}\right|-1$.
Suppose $m=m_{\min}^{\left(\left|\mathcal{Y}'\right|\right)}$ satisfies
eq. \ref{eq:target-sufficient-2-appendix}, we have 
\begin{eqnarray*}
 & \max_{\boldsymbol{x},\boldsymbol{x}'\in\mathcal{R}_{y_{a}}^{'}}\mathrm{d}\left(\boldsymbol{x},\boldsymbol{x}'\right)\le\min_{\boldsymbol{x}\in\mathcal{R}_{y_{a}}^{'},\boldsymbol{x}'\in\mathcal{R}_{y_{b}}^{'}}\mathrm{d}\left(\boldsymbol{x},\boldsymbol{x}'\right)\\
\text{where} & \mathcal{R}_{y_{a}}^{'}=\bigcap_{y\in\mathcal{Y}',y\ne y_{a}}\mathcal{R}_{y_{a},y}=\bigcap_{y\in\mathcal{Y},y\ne y_{a},y\ne y_{c}}\mathcal{R}_{y_{a},y} & ,\\
 & \mathcal{R}_{y_{b}}^{'}=\bigcap_{y\in\mathcal{Y}',y\ne y_{b}}\mathcal{R}_{y_{b},y}=\bigcap_{y\in\mathcal{Y},y\ne y_{b},y\ne y_{c}}\mathcal{R}_{y_{b},y}
\end{eqnarray*}

When $m$ is not changed and prototypes are not moved, 
\begin{eqnarray*}
\mathcal{R}_{y_{a}}=\bigcap_{y\in\mathcal{Y},y\ne y_{a}}\mathcal{R}_{y_{a},y} & \subseteq\bigcap_{y\in\mathcal{Y},y\ne y_{a},y\ne y_{c}}\mathcal{R}_{y_{a},y}=\mathcal{R}_{y_{a}}^{'}\\
\mathcal{R}_{y_{b}}=\bigcap_{y\in\mathcal{Y},y\ne y_{b}}\mathcal{R}_{y_{b},y} & \subseteq\bigcap_{y\in\mathcal{Y},y\ne y_{b},y\ne y_{c}}\mathcal{R}_{y_{b},y}=\mathcal{R}_{y_{b}}^{'}
\end{eqnarray*}
and 
\begin{eqnarray*}
\mathcal{R}_{y_{a}}\subseteq\mathcal{R}_{y_{a}}^{'} & \Rightarrow & \max_{\boldsymbol{x},\boldsymbol{x}'\in\mathcal{R}_{y_{a}}}\mathrm{d}\left(\boldsymbol{x},\boldsymbol{x}'\right)\le\max_{\boldsymbol{x},\boldsymbol{x}'\in\mathcal{R}_{y_{a}}^{'}}\mathrm{d}\left(\boldsymbol{x},\boldsymbol{x}'\right)\\
\mathcal{R}_{y_{a}}\subseteq\mathcal{R}_{y_{a}}^{'}\wedge\mathcal{R}_{y_{b}}\subseteq\mathcal{R}_{y_{b}}^{'} & \Rightarrow & \min_{\boldsymbol{x}\in\mathcal{R}_{y_{a}},\boldsymbol{x}'\in\mathcal{R}_{y_{b}}}\mathrm{d}\left(\boldsymbol{x},\boldsymbol{x}'\right)\ge\min_{\boldsymbol{x}\in\mathcal{R}_{y_{a}}^{'},\boldsymbol{x}'\in\mathcal{R}_{y_{b}}^{'}}\mathrm{d}\left(\boldsymbol{x},\boldsymbol{x}'\right)
\end{eqnarray*}

Thus, $\max_{\boldsymbol{x},\boldsymbol{x}'\in\mathcal{R}_{y_{a}}}\mathrm{d}\left(\boldsymbol{x},\boldsymbol{x}'\right)\le\min_{\boldsymbol{x}\in\mathcal{R}_{y_{a}},\boldsymbol{x}'\in\mathcal{R}_{y_{b}}}\mathrm{d}\left(\boldsymbol{x},\boldsymbol{x}'\right)$
is satisfied for any pair $\left(y_{a},y_{b}\right)$ where $y_{a},y_{b}\in\mathcal{Y}$,
even if we directly adopt $m=m_{\min}^{\left(\left|\mathcal{Y}'\right|\right)}$
when $\left|\mathcal{Y}\right|=\left|\mathcal{Y}'\right|+1$. So we
can conclude that $m_{\min}^{\left(\left|\mathcal{Y}'\right|\right)}\ge m_{\min}^{\left(\left|\mathcal{Y}'\right|+1\right)}$
is sufficient for $P_{2}$. By Lemma 3, $m_{\min}^{\left(2\right)}=2+\sqrt{3}$,
we can conclude that $m_{\min}^{\left(k\right)}=2+\sqrt{3},\forall k\ge3$
is sufficient for $P_{2}$.
\end{document}